\documentclass{article}

\usepackage{microtype}
\usepackage{graphicx}
\usepackage{booktabs} 
\usepackage{subcaption}
\usepackage{enumitem}
\usepackage{nicefrac}

\usepackage{hyperref}

\usepackage[accepted]{style2024}

\usepackage{amsmath}
\usepackage{amssymb}
\usepackage{mathtools}
\usepackage{amsthm}
\usepackage{xcolor}
\usepackage[most]{tcolorbox}

\usepackage[capitalize,noabbrev]{cleveref}
\usepackage{glossaries}
\usepackage{dsfont}
\usepackage{bm}

\newacronym{relu}{ReLU}{Rectifier Linear Function}
\newacronym{dnn}{DNN}{Deep Neural Network}
\newacronym{pi}{PI}{Private Inference}
\newacronym{sota}{SOTA}{state-of-the-art}
\newacronym{dp}{DP}{Differential Privacy}
\newacronym{ppml}{PPML}{Privacy Preserving Machine Learning}
\newacronym{nas}{NAS}{neural architecture search}
\newacronym{r18}{ResNet18}{ResNet18}
\newacronym{snl}{SNL}{Selective Network Linearization}
\newacronym{utk}{UTKFace}{UTKFace}
\newacronym{dr}{DR}{DeepReDuce}
\newacronym{sgd}{SGD}{Stochastic Gradient Descent}
\newacronym{kd}{KD}{Knowledge-Distillation}
\newacronym{nnl}{NNL}{Neural Network Linearization}

\newcommand{\losskd}{\mathcal{L}^{\text{KD}}}
\newcommand{\lr}{\eta}
\newcommand{\titlecustom}{Disparate Impact on Group Accuracy of Linearization for Private Inference}
\newcommand{\F}{\mathcal{F}}

\newcommand{\fnstar}{f_{n}^{\star}}

\newcommand{\x}{\boldsymbol{x}}
\newcommand{\R}{\mathbb{R}}

\newcommand{\paramsrelu}{\boldsymbol{\theta}}
\newcommand{\paramslin}{\boldsymbol{\tilde{\theta}}}
\newcommand{\paramslinr}{\boldsymbol{\tilde{\theta}^r}}

\newcommand{\aset}{\mathcal{A}}

\newcommand{\residual}[1]{R(#1)}
\newcommand{\gradient}[2]{\boldsymbol{g}_{#1}^{#2}}
\newcommand{\hessian}[2]{\boldsymbol{H}_{#1}^{#2}}
\newcommand{\calX}{\mathcal{X}}
\newcommand{\calY}{\mathcal{Y}}

\newcommand{\trainingsetshort}{S_T}

\newcommand{\testsetshort}{S_E}

\newcommand{\indicatorfunc}[1]{\mathds{1}\left[#1\right]}

\DeclareMathOperator*{\argmax}{arg\,\!max}
\DeclareMathOperator*{\argmin}{arg\,\!min}

\newcommand{\todocomment}[2]{\textcolor{red}{\textsuperscript{#1}: #2}}
\newcommand{\sdcomment}[1]{\textcolor{teal}{\textsuperscript{SD}: #1}}

\theoremstyle{plain}
\newtheorem{theorem}{Theorem}[section]
\newtheorem{proposition}[theorem]{Proposition}

\theoremstyle{definition}

\theoremstyle{remark}

\usepackage[textsize=tiny]{todonotes}

\titlerunning{\titlecustom}

\begin{document}

\twocolumn[
\title{\titlecustom}



\confsetsymbol{equal}{*}

\begin{authorlist}
\confauthor{Saswat Das}{equal,uva}
\confauthor{Marco Romanelli}{equal,nyu}
\confauthor{Ferdinando Fioretto}{uva}
\end{authorlist}

\confaffiliation{uva}{University of Virginia, Charlottesville, VA, USA}
\confaffiliation{nyu}{New York University, New York, NY, USA}

\confcorrespondingauthor{Saswat Das}{saswatdas@email.virginia.edu}
\confcorrespondingauthor{Marco Romanelli}{mr6852@nyu.edu}


\vskip 0.3in
]



\printAffiliationsAndNotice{\confEqualContribution} 

\begin{abstract}

Ensuring privacy-preserving inference on cryptographically secure data is a well-known computational challenge. To alleviate the bottleneck of costly cryptographic computations in non-linear activations, recent methods have suggested linearizing a targeted portion of these activations in neural networks. This technique results in significantly reduced runtimes with often negligible impacts on accuracy. In this paper, we demonstrate that such computational benefits may lead to increased fairness costs. Specifically, we find that reducing the number of ReLU activations disproportionately decreases the accuracy for minority groups compared to majority groups. To explain these observations, we provide a mathematical interpretation 
under restricted assumptions about the nature of the decision boundary, while also showing the prevalence of this problem across widely used datasets and architectures. Finally, we show how a simple procedure altering the fine-tuning step for linearized models can serve as an effective mitigation strategy.
\end{abstract}

\glsresetall

\section{Introduction}
\label{sec:introduction}
\emph{Private Inference} is the process of performing inference tasks on encrypted or private data, ensuring that the data remains confidential throughout the process. It has found application in ML settings where sensitive data is required for inference but should not be revealed to the model, for instance, when the model is owned by a cloud service provider and is not necessarily trusted.
Although promising, cryptographic computations are notoriously computationally expensive when applied to nonlinear functions \cite{MishraLSZP2020USENIX}, and, in the context of ML inference, \gls*{relu} activations are often identified as the primary cause of this computational burden \cite{ChoJRGH2022ICML,JhaGGR2021ICML,KunduLZLB2023ICLR}. 

To address this challenge, various approaches have proposed to approximate these non-linear activations using linear surrogates. The objective is to develop a new model that minimizes the number of \gls*{relu} activations while maintaining the highest possible accuracy. Over time, the proposed frameworks have evolved in sophistication---it has been demonstrated that not only the quantity of \gls*{relu} activations matters \cite{JhaGGR2021ICML}, but their placement within the network is also crucial \cite{ChoJRGH2022ICML}. 
Recent methods have advanced to the point where they can manage both the \gls*{relu} budget and the distribution of linearized neurons across the network, resulting in faster inference times with minimal sacrifice in accuracy \cite{KunduLZLB2023ICLR}. 

\begin{figure}[!tb]    
\centering
\includegraphics[width=0.95\columnwidth]{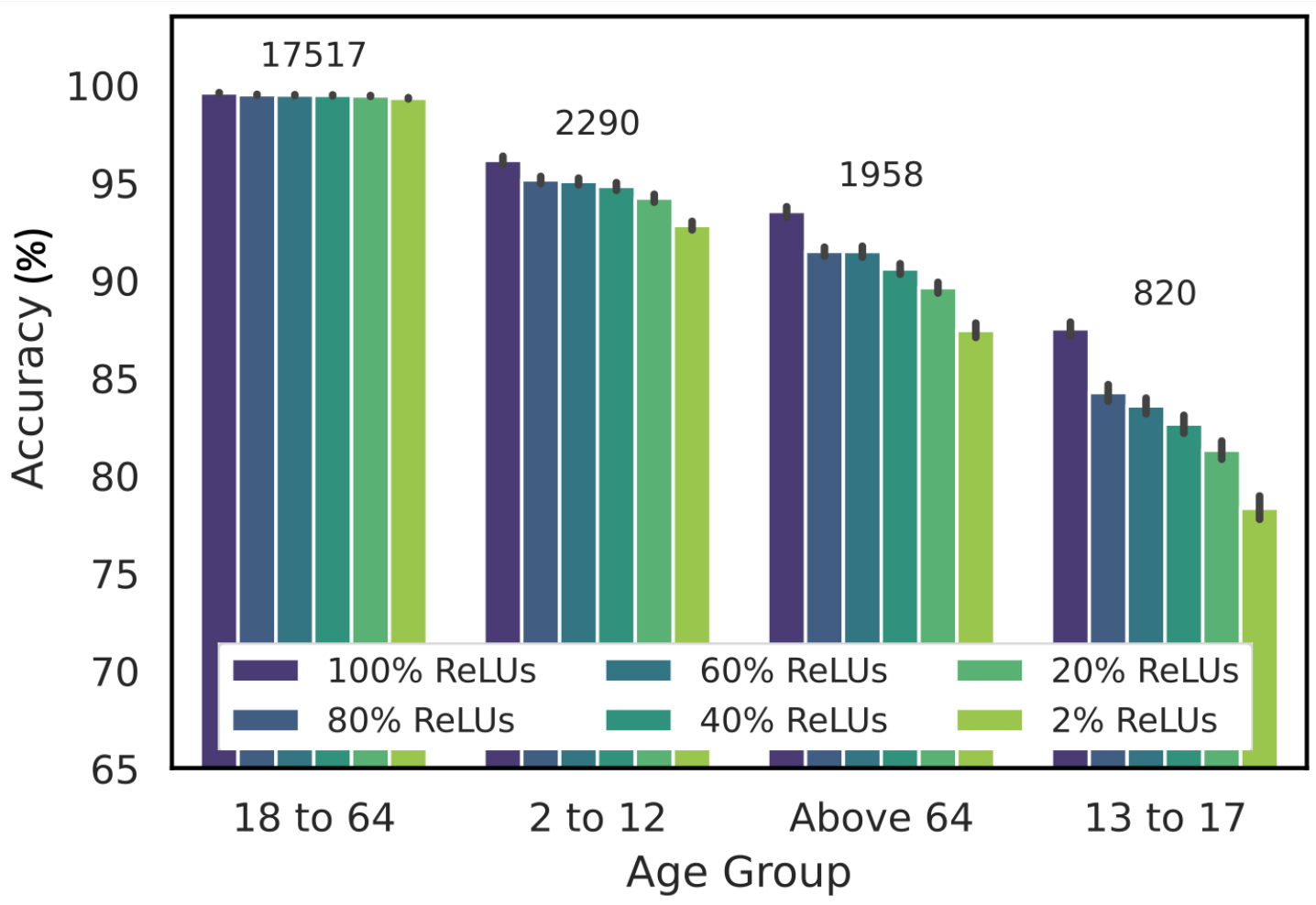}
\vspace{-6pt}
\caption{\acrlong*{r18} accuracy on \acrlong*{utk} across various ReLU linearization budgets using \acrshort*{snl}  \cite{ChoJRGH2022ICML}. 100\% ReLUs corresponds to the original model.
Subgroup sample sizes are shown above the corresponding bars. 
The accuracy for the majority group remains almost unchanged while other groups, in particular the minority one, are diversely impacted.
}
\label{fig:accuracy_assesment_intro}
\vskip -0.2in
\end{figure}

This paper builds on this body of work and observes that while these methods are indeed effective at trading run-time reduction for marginal global accuracy losses, these accuracy decreases are unevenly distributed among different subgroups. We find that the accuracy loss is more pronounced for underrepresented subgroups and that this impact intensifies as more ReLUs are linearized. 
This effect is depicted in \cref{fig:accuracy_assesment_intro}, which shows the accuracy impact across age groups on a facial recognition task. Remarkably, while the majority group (left-most) remains relatively unharmed, minorities exhibit considerable reduction. 
\emph{These results are far from ``expected''}: they challenge the assumption that linearization uniformly affects the learning model and underscore the need for different approaches in model optimization, particularly in diverse datasets.

\textbf{Contributions.}
This paper makes the following contributions:
\textbf{(1)} It observes, for the first time, a Matthew effect on the reduction of precision caused by the targeted linearization process adopted to reduce the number of \gls*{relu} functions in private inference methods. 
Further, it shows that such disparity is exacerbated as the number of approximated \gls*{relu} functions grows.
\textbf{(2)} Next, it presents a mathematical interpretation of this phenomenon, which relates these disparity effects with the approximation capabilities of \gls*{relu} functions  
when assumptions on the space of the decision boundaries can be made.
\textbf{(3)} It further shows that the effect of \gls*{relu} reduction on fairness is vastly present for commonly deployed algorithms and models when trained on unbalanced datasets. 
\textbf{(4)} Finally, the paper proposes a simple yet effective mitigation strategy that can be applied to any framework for \gls*{relu} reduction during the fine-tuning phase, showing favorable results in terms of fairness and overall accuracy reduction. 

\section{Related Work}
\label{sec:relatedwork}
\textbf{Frameworks for ML Private Inference.}
The relationship between presence of \gls*{relu} activations and the inference time of a private inference model was first investigated by \citet{MishraLSZP2020USENIX}. The seminal solution proposed in this work is based on automatic generation of neural network architecture configurations, that navigates the performance-accuracy trade-offs over a given cryptographic protocol. Building on this intuition, \citet{GhodsiVRG2020NeurIPS,LouSJJ2021ICLR} propose to selectively replace \gls*{relu} functions with, computationally more efficient, polynomial operations. \gls*{dr} \cite{JhaGGR2021ICML} takes this approach a step further by proposing a multi-step optimization, where portions of the network are replaced with linear functions, and then finetuned to recover the accuracy of the original model. The main drawback of these methods is the limited control over the ``distribution'' of the \gls*{relu} throughout the net. To tackle this issue, \gls*{snl} \cite{ChoJRGH2022ICML} proposes a custom activation node, where the linear and rectified behavior are parametrized and learned during training through a parametric mask. In line with this paradigm, the approach in \citet{KunduLZLB2023ICLR} introduces a novel measure of non-linearity sensitivity that helps reduce the need for manual efforts in identifying the best \gls*{relu} budget and placement.

\textbf{Piece-wise Approximation with \gls*{relu} Functions.}
In addition to observing disparate impacts from methods that reduce the number of ReLU functions in neural networks, our work seeks to explain these effects by linking them to the expressiveness and approximation capabilities of a \gls*{dnn}. These studies trace back to seminal works in \citet{Cybenko1989MCSS,Hornik1991NN}. The widespread use of \gls*{relu} activations to introduce non-linearity in the learning of decision boundaries has motivated early studies about their approximation power \cite{MontufarPCB2014NeurIPS,PanS2016ICML}. Several papers have observed the effect of \gls*{relu} activations in the learning process \cite{HaninR2019NeurIPS,LinJ2018NeurIPS,LuPWHW2017NeurIPS,ShenYZ2021JMPA,Yarotsky2017NN}. Among others, \citet{AroraBMM2018ICLR,LiuL2021Neurocomputing} distinguish themselves for contributions to the field, analyzing the connection between the number of \gls*{relu} activations and that of the approximated piecewise functions, along with a \gls*{relu}-dependent bound on the approximation error.

\textbf{Privacy and Fairness Tradeoff in Machine Learning.}
Among other works on the tradeoff 
between privacy and fairness in machine learning, 
\citet{BagdasaryanPS2019NeurIPS} 
provides mostly empirical evidence of 
the disparate effect of differentially private training algorithms on 
fairness, while \citet{FarrandMST2020PPMLws} further 
investigates this phenomenon by considering different 
levels of differential privacy and unbalance in the datasets. These works provide empirical evidence to 
show the noise injection 
and
the gradient clipping involved in 
differentially private Stochastic Gradient Descent 
impact the overall accuracy, 
and especially that of underrepresented 
classes.
The privacy framework we consider is extremely different, 
as the diverse impact is not a direct consequence 
of the privacy preservation framework design, 
but of the architectural adjustments made to 
speed up cryptographic secure inference 
.
Such adjustments impact the 
approximation 
capabilities of the model 
resulting in disparate 
accuracy across different groups.

Finally, model compression has also been shown to induce fairness issues. Empirical observations reported  that quantization, network compression, and knowledge distillation could amplify the unfairness in different learning tasks \cite{lukasik2022teachers,Hooker2020WhatDC,Hooker2020CharacterisingBI,Joseph2020GoingBC,Blakeney2021SimonSE,Ahn2022WhyKD}.
Inspired by these works, we  
show how \gls*{relu} linearization techniques may adversely affect the fairness of the resulting predictors. 

\section{Settings}
\label{sec:settings}
The paper considers a dataset $\trainingsetshort$ consisting of $N$ individual data points $(\bm{x}_i, a_i, y_i)$, 
with $i\in[N]$ drawn i.i.d.~from an unknown distribution $\Pi$. Therein, $\bm{x}_i \in {\cal X}$ is a 
feature vector, $a_i \in {\cal A} = [M]$, for some finite $M$, is a demographic group attribute, 
and $y_i \in {\cal Y}$ is a C-class label (${\cal Y} = [C]$). 
The goal is to learn a classifier $f_{\boldsymbol{\theta}}: \calX \to \calY$, where $\boldsymbol{\theta}$ is a vector of real-valued 
parameters. The classifier $f$ is often defined through a \emph{soft classifier} $h_{\boldsymbol{\theta}} : \calX \to \mathbb{R}^C$, that, for a given input $\bm{x}$, produces scores for each label in $\calY$, defining $f_{\boldsymbol{\theta}}(\bm{x}) = \argmax_j h_{\boldsymbol{\theta}}(\bm{x}; j)$ where
$h_{\boldsymbol{\theta}}(\bm{x}; j)$ denotes the $j$-th component of $h_{\boldsymbol{\theta}}(\bm{x})$.
The model quality is measured in terms of a loss function $\ell:\calY \times \calY \to \mathbb{R}_+$ and the training minimizes the empirical risk function $J(_{\boldsymbol{\theta}}; \trainingsetshort)$: 
\[
    {\boldsymbol{\theta}}^* = \argmin_{{\boldsymbol{\theta}}} J({\boldsymbol{\theta}}; \trainingsetshort) 
     = \frac{1}{|S_T|} \sum_{(\bm{x},a, y) \in \trainingsetshort} \ell\left( h_{\boldsymbol{\theta}}(\bm{x}), y\right).
\]
The paper studies the disparate impacts of classifiers $f_{\boldsymbol{\theta}}$ when it is subject to a ReLU partial reduction. 
The fairness notion employed is that of \emph{accuracy parity} \cite{barocas-hardt-narayanan}, which holds when the classifier's misclassification rate is conditionally independent of the protected group. That is, for any $\bar{a} \in {\cal A}$
\[
    \Pr\left(f_{\boldsymbol{\theta}}(\bm{x}) \neq y | a = \bar{a}\right) = \Pr\left(f_{\boldsymbol{\theta}}(\bm{x}) \neq y \right).
\]
In other words, this property advocates for equal errors of the classifier on different subgroups of inputs. Empirically, it is measured by comparing the accuracy rates over an evaluation set $\testsetshort$ comprised of data points drawn from the same distribution $\Pi$ as the training set.

\section{Why Network Linearization May Increase Unfairness?}
\label{sec:problemstatement}
This section introduces theoretical insights to elucidate the causes behind the observed disparity effects resulting from the linearization of ReLU reduction. 
We start by recalling some fundamental results that will help us develop our analysis in the rest of the paper. The first results \cite{LiuL2021Neurocomputing} defines a connection between the number of piecewise functions and the approximation error of convex functions. 
Throughout the section, ${\cal F}$ denotes the set of strict convex univariate functions. 
\begin{proposition}
    \label{prop:approximation_error}
    Consider a function $f\in \F$ and let $\fnstar$ be its optimal approximation through a piecewise linear function with $n$ segments. Then, the approximation error $\Delta(
    \fnstar)$ is bounded by the number of its segments, and it decreases at a rate of $O(\frac{1}{n^{2}})$.
\end{proposition}
The second result \cite{AroraBMM2018ICLR} upper-bounds the number of pieacewise linear functions that can be represented by a ReLU network.
\begin{proposition}
    \label{prop:relu_upper_bound}
    Given a \gls*{relu} $\R\rightarrow\R$ \gls*{dnn} with $k$ hidden layers, and layer widths $\omega_{1},\dots,\omega_{k}$, the number of attainable linear pieces is at most $2^{k-1}\cdot(\omega_{1}+1)\cdot\omega_2\cdot\dots\cdot\omega_k$.
\end{proposition}
This bound links the network's capability to approximate complex decision boundaries with the count of \gls*{relu} functions it contains (see \cref{sec:example-prop4.2} for an illustrative example). Together with the previous result on the approximation error linked to a specific number of pieces  (\cref{prop:approximation_error}), it sets the stage for our subsequent analysis. We will illustrate that linearizing a number of ReLU functions not only leads to a potential decrease in accuracy when approximating nonlinear boundaries but it may also result in varying degrees of accuracy impact across distinct groups.

\begin{figure}[!tb]
 \centering
     \begin{subfigure}{.7\columnwidth}
            \centering
            \includegraphics[width=\textwidth,trim={0 0 0 1.8in},clip]{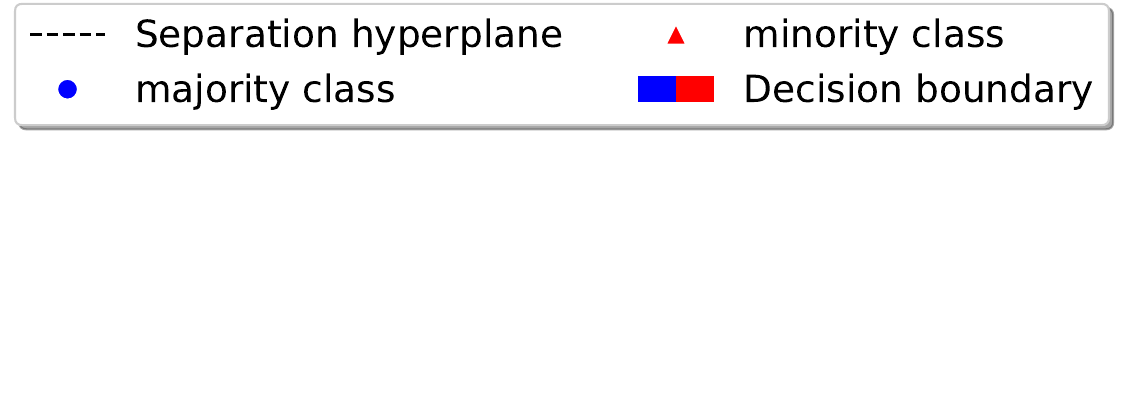}
       \end{subfigure}
       \vskip -0.5in
         \begin{subfigure}[b]{.47\columnwidth}
             \centering
             \includegraphics[width=\textwidth,trim={0.5in 0.5in 0.5in 0.5in},clip]{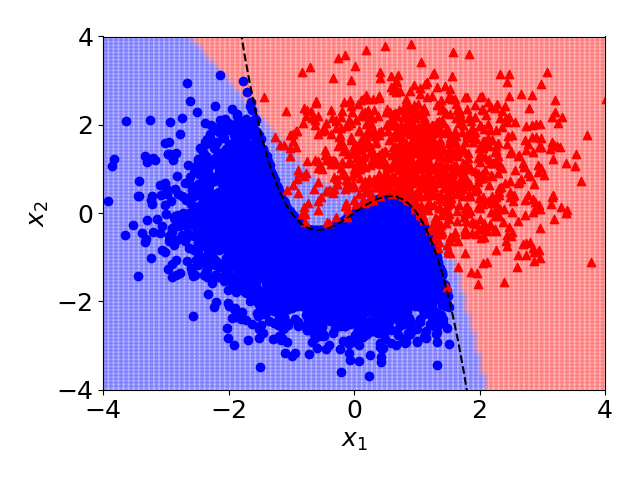}
             \vskip -0.05in
             \caption{Original \gls*{relu} model.}
             \label{fig:toy1}
        \end{subfigure}
        \hfill
        \begin{subfigure}[b]{.47\columnwidth}
             \centering
             \includegraphics[width=\textwidth,trim={0.5in 0.5in 0.5in 0.5in},clip]{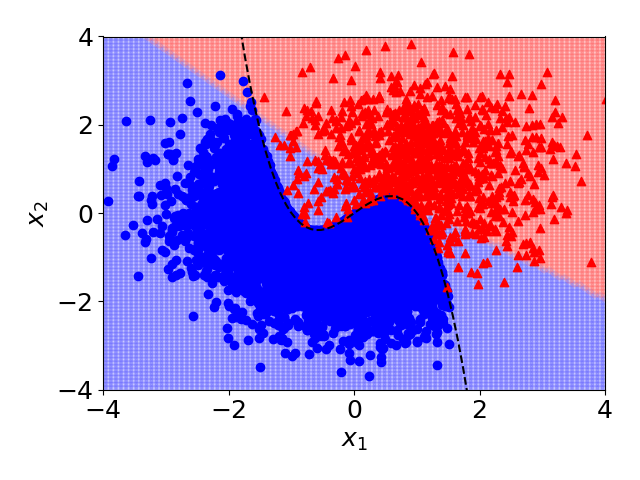}
             \vskip -0.05in
             \caption{$50\%$ ReLU linearization.}
             \label{fig:toy2}
         \end{subfigure}
        \vskip -0.1in
        \caption{Decision boundary estimated by a \gls*{relu} model (left) and a linearized model (right). The ground truth decision boundary (black dotted line) separates majority (blue) from minority (red) samples. The linearized model shows an inferior approximation of the decision boundary, and therefore a lower prediction accuracy. Moreover, its decision boundary is deeper into to the minority class resulting in accuracy disparities.
        }
        \label{fig:toy}
\end{figure}

We start with an illustrative experiment, detailed in \cref{fig:toy}. It analyzes a dataset requiring a non-linear decision boundary for class separation (black dotted line), representing a protected group in fairness terms, with majority (blue) and minority (red) classes. We compare two \gls*{dnn} models with identical structures but different activation functions: a \gls*{relu} model and a modified model where half of the \gls*{relu} activations are replaced with linear activations, as shown in \cref{fig:toy1} and \cref{fig:toy2}, respectively. Notice that, after training both models from scratch, the \gls*{relu} model achieves better boundary approximation and accuracy (99.3\% globally, 100.0\% for the majority class, and 93.0\% for the minority class), while the modified model shows comparable global accuracy (99.0\%) but much lower fairness, indicated by reduced accuracy for the minority class (88.0\%) compared to the original \gls*{relu} model.

\subsection{Decision Boundary Approximation and Fairness}
These insights indicate that there exists classification instances where the quantity of \gls*{relu} functions within a \gls*{dnn} enhances the approximation of decision boundaries, thereby improving prediction accuracy. 
Although empirical results often apply to much more complex settings than those which can be analyzed formally, they align with the forthcoming theoretical findings. 
In the following, we will use $\paramsrelu$ to denote the parameters of the original ReLU network, which is assumed to have $R$ \gls*{relu} functions, and $\paramslinr$ those associated to the linearized model, which retains $r$ ReLU functions. 
We omit the superscript when it is sufficient to denote a network retaining $1\leq r < R$ \gls*{relu} functions.

\begin{proposition}
    Let $f \in \F$ be the optimal decision boundary for a classification task, and assume that $\paramsrelu$ implements $f_n^\star$, i.e., the minimal error approximation. 
    For a sample set $S^a = \{(\bm{x}, \bar{a}, y) \sim \Pi \vert \bar{a} = a\}$ with data points drawn from $\Pi$ containing exclusively members of group $a \in {\cal A}$, it holds 
    \[
        \residual{a} = J(\paramslin; S^a) - J(\paramsrelu; S^a) \geq 0,
    \] 
    for any $a \in {\cal A}$,
    where all the models are assumed to be trained to achieve their best possible performance.
\end{proposition}
The above follows directly from the optimality of $\bm{\theta}$. In other words, for this class of boundaries, any degree of ReLU linearization results in models with higher loss, within the class of decision boundaries considered. 

\begin{figure}
    \centering
    \includegraphics[width=0.75\linewidth,trim={0.0in 0.11in 0.0in 0.0in},clip]{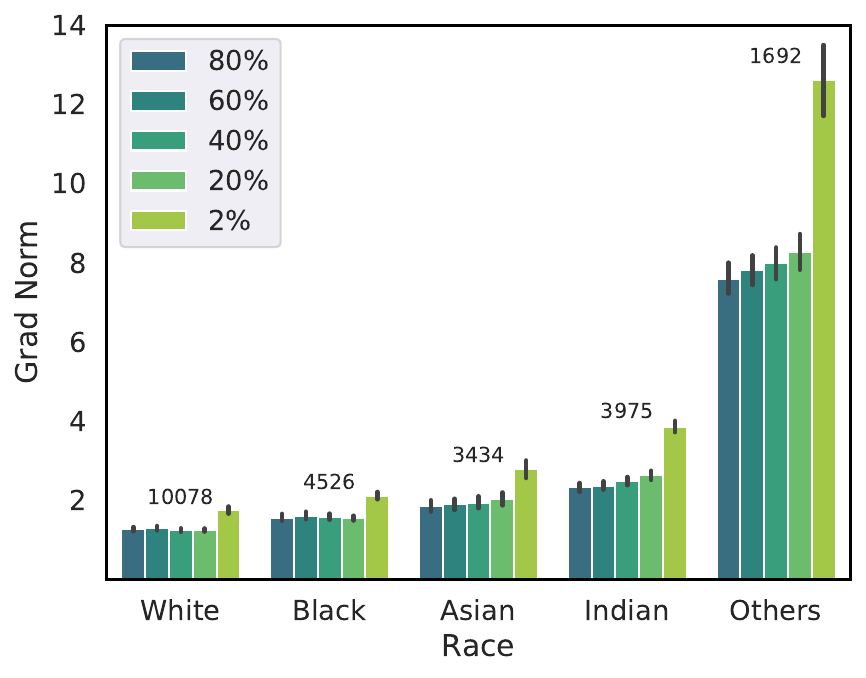}
    \caption{Grad Norms for classification on UTKFace with race labels using ResNet18 and SNL-based \cite{ChoJRGH2022ICML} \gls*{relu} linearization across various \gls*{relu} budgets. Subgroup sizes are reported on top of each bar group.}
    \label{fig:gradnorms}
\end{figure}

The next result is adapted from \citet{TMF:neurips21} which exploits a similar approximation in the context of differentially private stochastic gradient descent. We upper bound the drop in loss for a given group $a\in {\cal A}$ due to \gls*{relu} linearization by two key interpretable components: the \emph{gradient norm} of the samples in group $a$, 
and the \emph{maximum eigenvalues of the Hessian} of the loss function associated with such a group. 
\begin{theorem}
    \label{prop:bound_residual_loss}
    For a sample set $S^a = \{(\bm{x}, \bar{a}, y) \sim \Pi \vert \bar{a} = a\}$ with data points drawn from $\Pi$ containing exclusively members of group $a \in {\cal A}$, the difference between the risk functions of some protected group $a$ of a model $\paramsrelu$ trained on a ReLU network and one $\paramslin$ trained on a linearized ReLU network, is bounded by:
    \begin{align}
        \label{eq:bound_residual_loss}
        \residual{a} \leq &\left\|\gradient{a}{} \right\| \times\|\paramslin - \paramsrelu\|+ 
        \frac{1}{2} \lambda\left(\hessian{a}{} \right) \times \|\paramslin - \paramsrelu\|^2
        \nonumber\\
        & + O\left(\|\paramslin - \paramsrelu\|^3\right),
    \end{align}
    where $\bm{g}_a = \nabla J(\paramsrelu; S_T^a)$ describes the gradient norm of samples in group $a$, ${\bm{H}^{\ell}_{a} = \nabla^2 J(\paramsrelu; S_T^a)}$ is the Hessian of the loss function associated with group $a$, and $\lambda(\Sigma)$ denotes the maximum eigenvalue of matrix $\Sigma$.
\end{theorem}
The proof is based on the Taylor expansion of the loss function around the parameters $\paramsrelu$ and is relegated to \cref{app:proof_1}. In the above, and throughout the paper, the gradients are understood to be over the model's parameters. 

There are two major takeaways from \cref{prop:relu_upper_bound}. For given parameter settings $\paramsrelu$ and $\paramslin$, the first component of the bound is influenced by the gradient norm of the loss function for each group $a\in\aset$. Informally, the gradient provide information on how close the model is to local optima; larger gradients suggest sub-optimality, while smaller ones indicate proximity to these optima. Typically, in converged models, underrepresented groups tend to have larger gradient norms, implying a greater distance from local optima (cf.~\cref{subsec:underrepresented_groups_have_larger_gradients}). This aspect is highlighted in \cref{fig:gradnorms}, where each shade represents a linearized model variant with varying counts of retained \gls*{relu} functions. Here, it's evident that gradient norms are higher for underrepresented groups, decreasing as more \gls*{relu} functions are retained.
Next, the bound's second term is governed by the Hessian of the loss function for each group $a\in\aset$. The Hessian reflects the local curvature, or sharpness, of the loss function. Generally, smaller Hessian eigenvalues suggest (with certain caveats) a flatter curve for the group loss, potentially indicating better generalizability to unseen samples.

\begin{figure}
     \centering
     \includegraphics[width=0.75\linewidth]{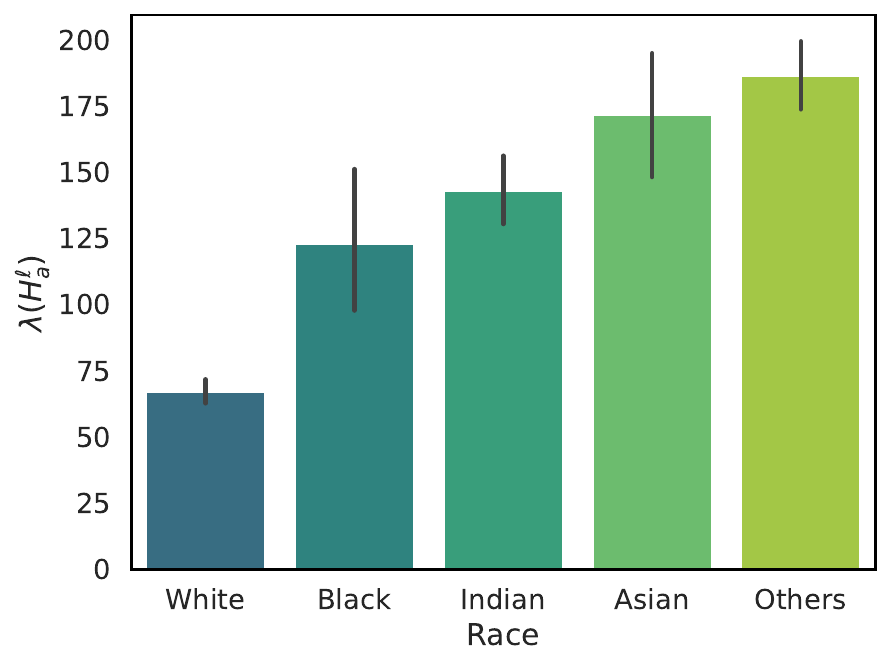}
        \caption{ResNet-18: Highest eigenvalues of Hessians for UTKFace with race labels obtained from the base model.}
        \label{fig:UTKFace_Hessian_Plots}
\end{figure}

\begin{proposition}
    \label{prop:max_eig}
    Consider a binary classifier $f_{\paramsrelu}$ trained using binary cross entropy loss. For any group $a\in\aset$ the maximum eigenvalue of the group Hessian $\lambda\left(\hessian{a}{}\right)$ can be upper bounded by
    \begin{equation}
        \label{eq:bound_on_group_hessian}
        \lambda\left(\hessian{a}{\ell}\right) \leq Z_1 + Z_2,
    \end{equation}
    where,
    \begin{align*}
        Z_1 &= \frac{1}{\left|S^{a}\right|}\!\sum_{(\bm{x}, a, y) \in S^a} \underbrace{h_{\paramsrelu}(\x)\left(1-h_{\paramsrelu}(\x)\right)}_{\text {Proximity to decision boundary }} \!\!\!\!\!\times\left\|\nabla h_{\paramsrelu}(\x)\right\|^2,\\
        Z_2&=\underbrace{\left|f_{\paramsrelu}(\x)-y\right|}_{\text {Error }} \times \lambda\left(\nabla^2 h_{\paramsrelu}(\x)\right).
    \end{align*}
\end{proposition}
The proof (see \cref{app:proof_1}) relies on derivations of the Hessian associated with the model loss function and Weyl inequality with the notion of proximity to decision boundary is derived by \cite{cohen2019certified}. 
As a consequence, groups with small Hessians eigenvalues (those generally distant from the decision boundary and highly accurate) tend to be less sensitive to the effects of the ReLU reduction rate. Conversely, groups with large Hessians eigenvalues tend to be affected by the ReLU reduction rate to a greater extent, typically resulting in larger excessive losses. 

\cref{fig:UTKFace_Hessian_Plots} illustrates the theoretical insights presented in \cref{prop:max_eig}: crucially, the value of $\lambda\left(\hessian{a}{\ell}\right)$ corresponding to the underrepresented groups are consistently higher than those  corresponding to highly represented groups.

$\blacksquare$ {\bf Remark about \cref{prop:bound_residual_loss}.} 
It is important to note that the bound presented in \cref{prop:bound_residual_loss} also depends on the distance $| \paramslin - \paramsrelu|$ between the ReLU-linearized model and the original model. Generally, a formal relationship between this distance and the rate of ReLU linearization is challenging to establish, as it inherently depends on the specific model and the distribution of ReLUs induced by the adopted linearization scheme. 

Empirically, our experiments consistently observe this result, as shown in \cref{fig:TestAccsandOtherMetrics} (left): the distance $| \paramslin - \paramsrelu|$ decreases with the increase of the ReLU budget. It is important to note that the model parameters retain information about the activation functions and the data split $S_a$ used during the training phase when the gradients of the loss w.r.t. the parameters are computed. Thus, a weight vector $\paramslin$ may induce different outputs based on where in the network the various ReLU activations are preserved.  
Additionally, while our results are based on networks that have been extensively trained, possibly to or very close to convergence, certain corner cases could affect our analysis of the bound. For instance, the bound may not hold for poorly trained networks or if we were to impose equal weights on linearized and non linearized models. In that case the distance $\| \paramslin - \paramsrelu\|$ would reduce the bound to zero, while the left-hand side of the bound would still be positive by construction. 

To be more precise, the function $J(\cdot)$ should be considered not solely as a function of the model’s parameters, but also of the data, and the set of activation functions. This would lead to a more complex bound that increases with the distance between the \emph{channels}, i.e., the posterior probabilities induced by $h_{\paramsrelu}(\cdot)$ and $h_{\paramslin}(\cdot)$ on $S_a$. While this refinement could render our analysis more precise, it would also introduce much notation, while also not altering our key observation that \emph{linearizing models may have disparate impacts on underrepresented groups}.
\hfill$\blacksquare$



\section{Empirical Results on the Fairness Analysis}
\label{sec:empiricalresults}
We next present the key findings from our empirical analysis on how \gls*{relu} linearization affects the fairness of linearized models. These findings build upon and extend the theoretical insights from earlier sections to scenarios involving high-dimensional and non-convex decision boundaries.
In summary, we show that {\bf (1)} model linearization produces disparate degradation of performance across groups and that such disparity is enhanced with the larger amounts of linearization; {\bf (2)} the relative magnitude of gradient norms for a group serves as a reliable indicator of the group's accuracy decline resulting from ReLU linearization; {\bf (3)} groups located farther from (or closer to) the decision boundary are less (more) likely to experience accuracy reduction at varying levels of ReLU linearization\footnote{Code:
\url{https://github.com/SaswatD27/ICML_Linearization_Disparate_Impact}}.

\textbf{Datasets and Models.}
We adopt three datasets:
\begin{itemize}[leftmargin=*, parsep=0pt, itemsep=0pt, topsep=0pt]
    \item \textbf{UTKFace} \cite{zhifei2017cvpr}. Containing over 20,000 face images with annotations for age, gender, and race is widely used for computer vision tasks. Our experiments use \textbf{age} and \textbf{race} as protected groups.
    \item \textbf{SVHN} (Digits) \cite{SVHN}. Containing $60,000$ $32\times32$ RGB images of digits of house number plates.
    \item \textbf{CIFAR-10} \cite{krizhevsky2009learning}. Containing $60,000$ $32\!\times\!32$ RGB images of 10 classes (airplanes, cars, birds, cats, deer, dogs, frogs, horses, ships, and trucks).
\end{itemize}

To assess the effect of network linearization on groups with various representations, we evaluate several models and architectures on unbalanced datasets. While UTKFace and SVHN are naturally unbalanced, we unbalanced CIFAR-10 by selecting 90\% of the samples for each of 5 classes, and retaining 10\% of the samples for each of the remaining classes. All reported metrics are average over 10 random seeds. 
The UTKFace dataset includes images with a large variation in age, ranging from 0 to 116 years old. As customary in demography, the ages are categorized into four groups: 2-12 (young children), 13-17 (adolescents), 18-64 (working-age individuals), and over 64 (senior citizens) for the experiments in this paper and these groups contain, respectively, $10.14\%$, $3.63\%$, $77.56\%$, and $8.67\%$ of the total samples in the dataset. Data points corresponding to ages 0 and 1 are considered outliers and thus omitted for this task.
Race classification on UTKFace utilizes the dataset's provided labels: white, black, Asian, native Indian, and others which correspond to $42.51\%$, $19.09\%$, $14.49\%$, $16.77\%$, and $7.14\%$  of the total samples in the dataset, respectively.
Finally, digit recognition studies are conducted on the naturally imbalanced SVHN dataset. This dataset contains 10 digits from 0 to 9 which have  
$6.75\%$, $18.92\%$, $14.45\%$, $11.60\%$, $10.18\%$, $9.39\%$, $7.82\%$, $7.64\%$, $6.89\%$, and $6.36\%$
of the total samples in the dataset, respectively

In this analysis two popular architectures are considered: ResNet18 and ResNet34 \cite{He2015DeepRL}. These architectures are the main ones used in the linearization frameworks of \cite{JhaGGR2021ICML,ChoJRGH2022ICML}.

\begin{figure}[!tb]
    \centering
    \begin{subfigure}{0.23\textwidth}
        \includegraphics[width=\textwidth]{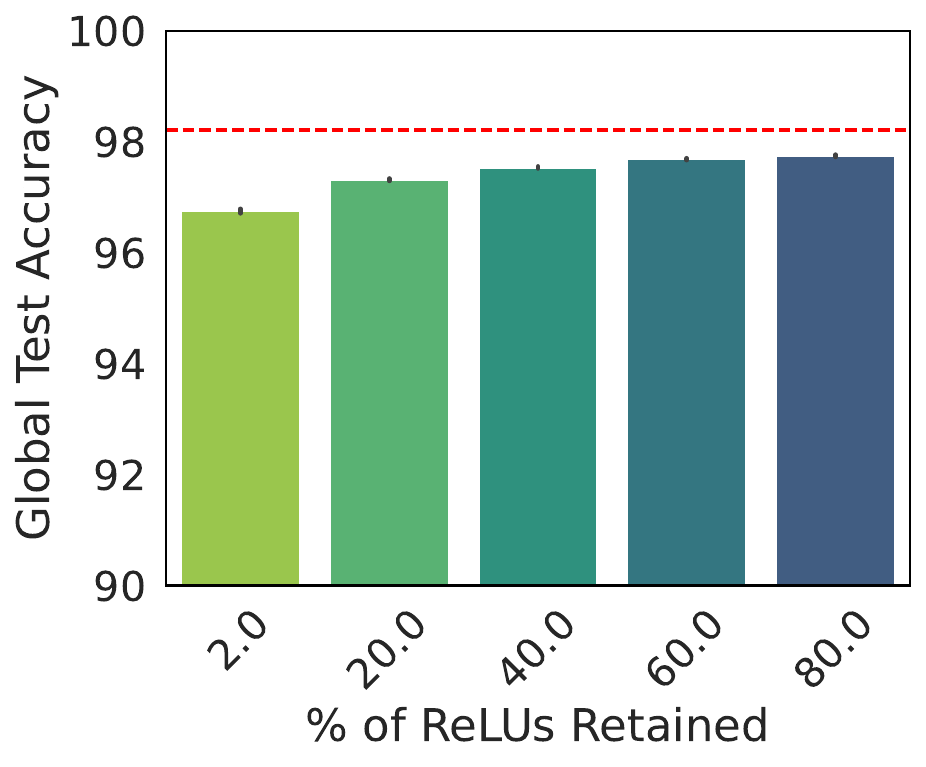}
        \caption{\gls*{snl} linearization}
        \label{fig:snl_glob_acc_utk}
    \end{subfigure}
    \hfill
    \begin{subfigure}{0.23\textwidth}
        \includegraphics[width=\textwidth]{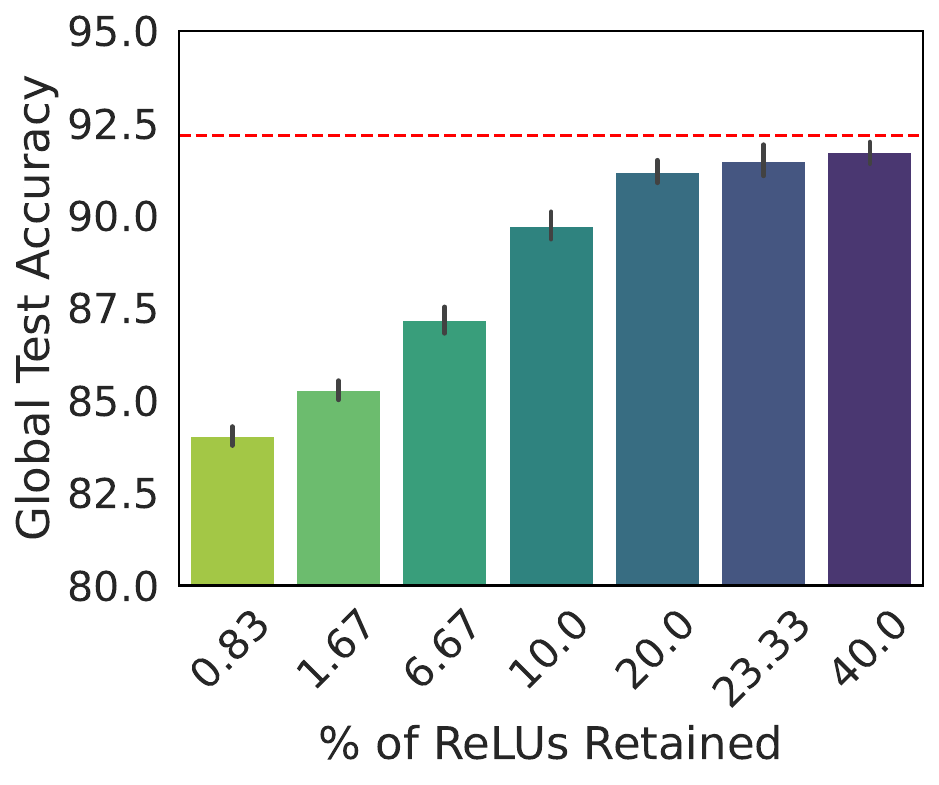}
        \caption{\gls*{dr} linearization}
        \label{fig:dr_glob_acc_utk}
    \end{subfigure}
    \caption{Global test accuracy on UTKFace with age labels using ResNet18 for \gls*{snl} and \gls*{dr}. The red horizontal line represents the original (no \gls*{relu} linearization) global test accuracy.}%
    \label{fig:GlobalTestAccs_UTKFace_age_SNL_RN34}
\end{figure}

\begin{figure*}[!t]
    \centering
    \includegraphics[width=0.23\linewidth]{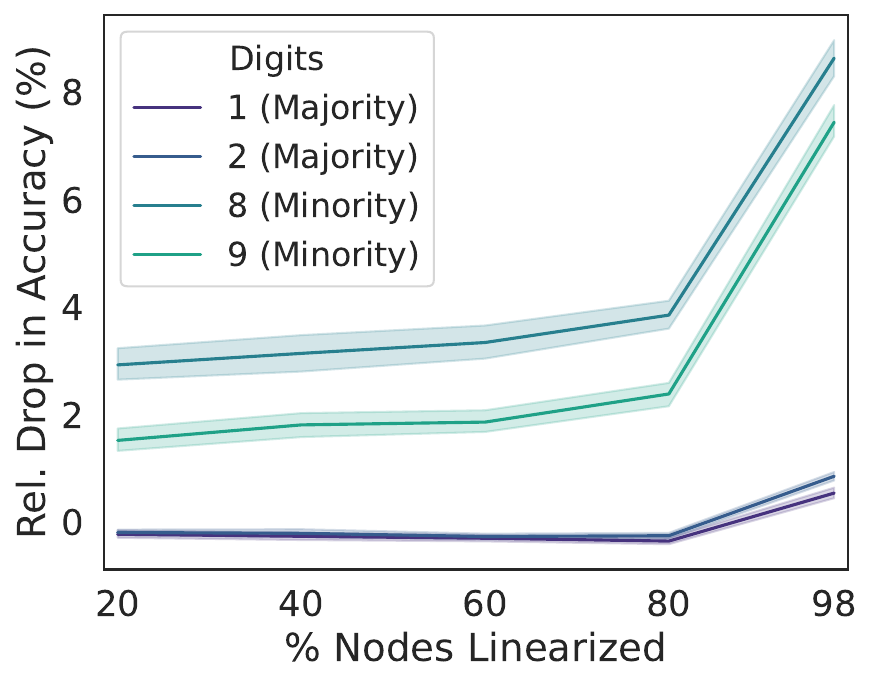} 
    \includegraphics[width=0.23\linewidth]{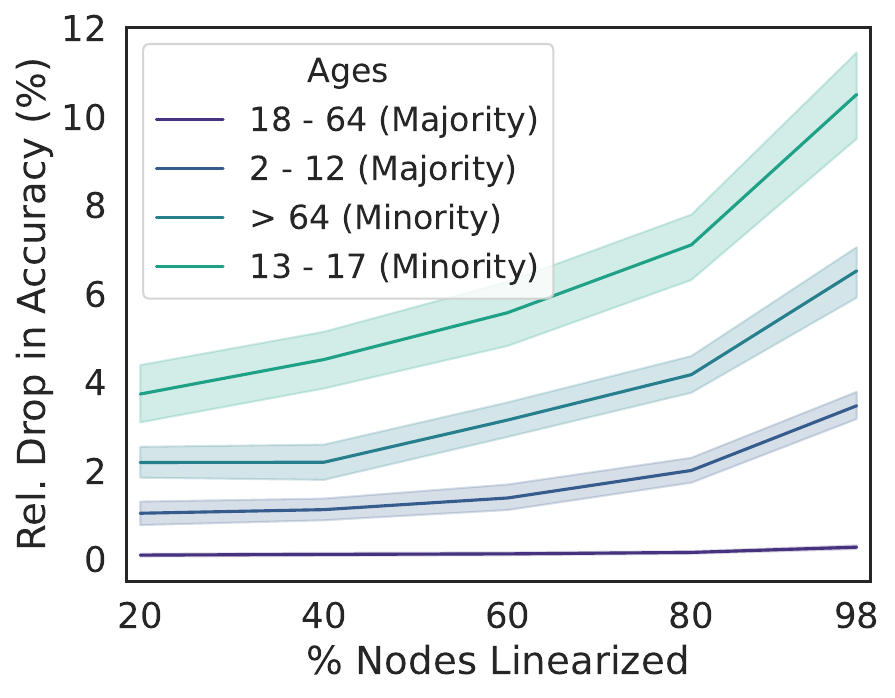} 
    \includegraphics[width=0.23\linewidth]{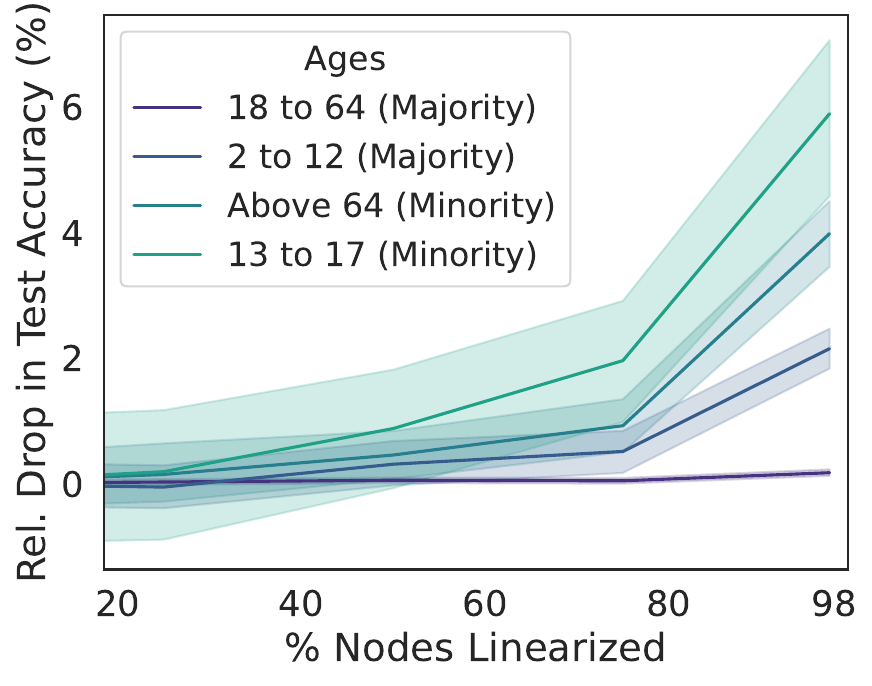}
    \includegraphics[width=0.24\linewidth,trim={0 0 0 0.1in},clip]
        {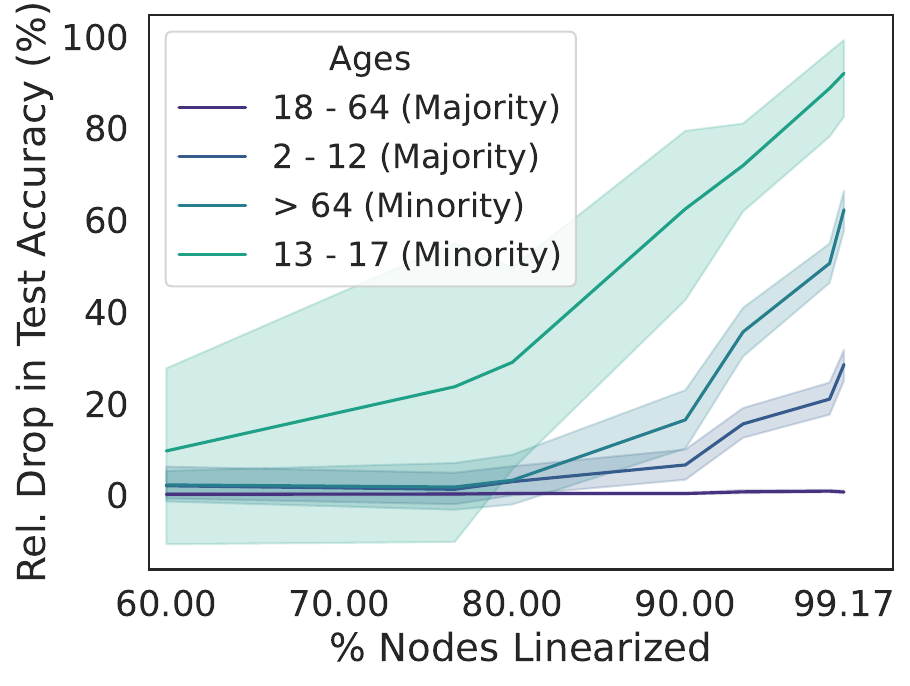}\\ 
    \includegraphics[width=0.23\linewidth]
    {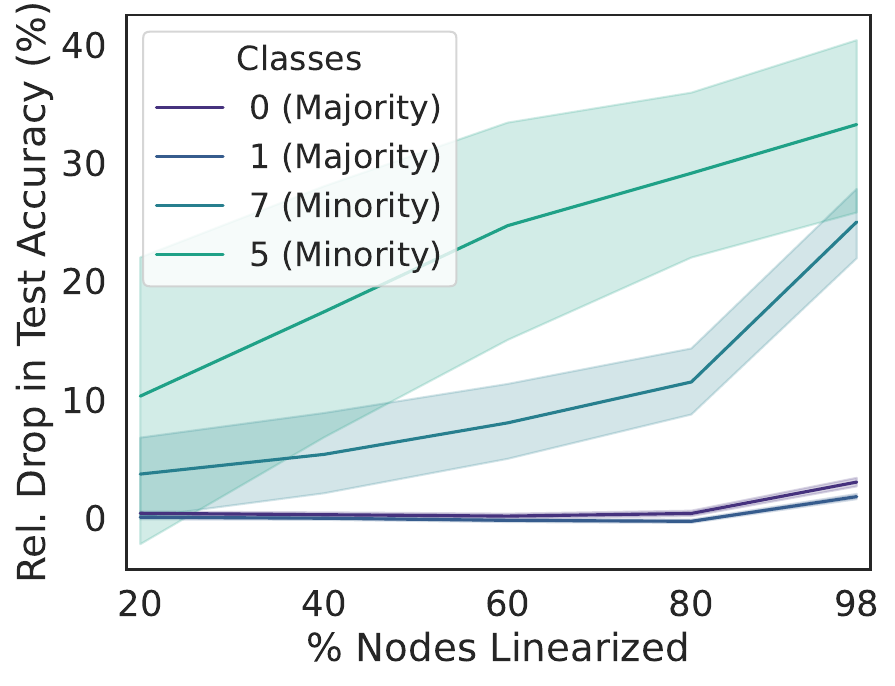} 
    \includegraphics[width=0.23\linewidth]{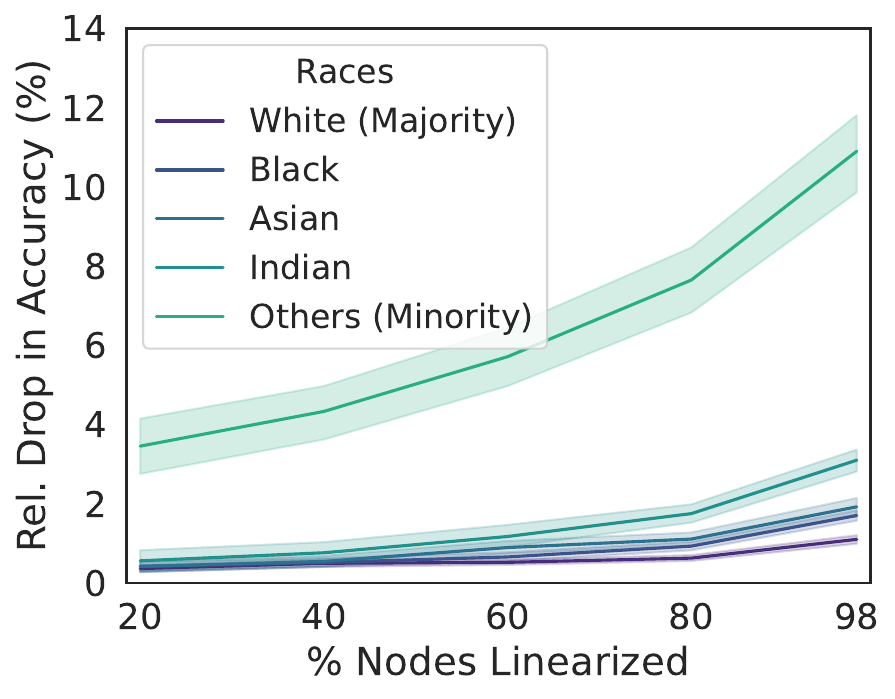}  
    \includegraphics[width=0.23\linewidth]{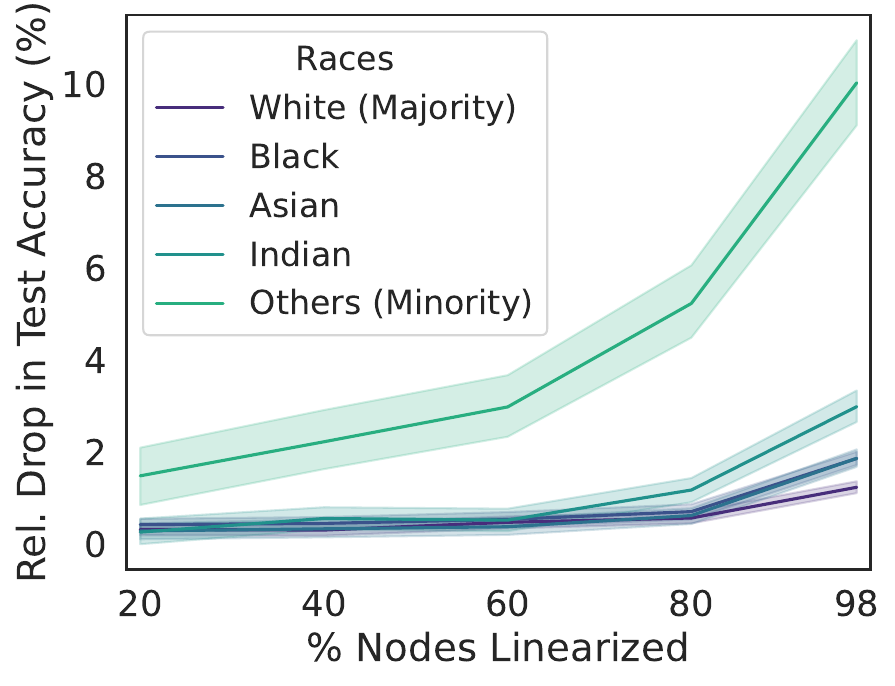} 
    \includegraphics[width=0.24\linewidth]{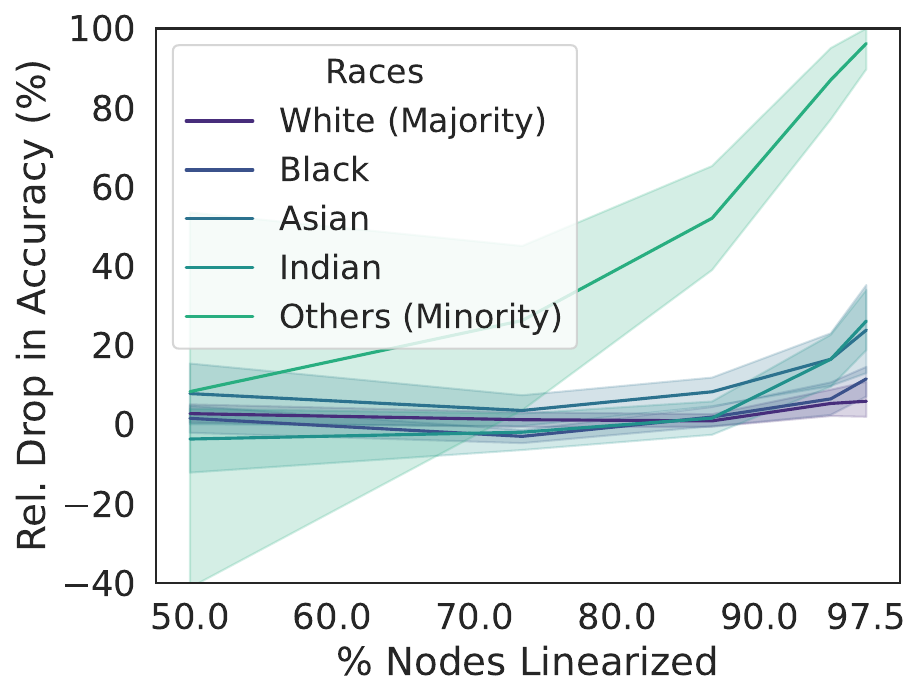}
    \caption{\gls*{snl}: Relative test accuracy drop for different datasets , models, and \gls*{relu} linearization methods.
    \textbf{1st col:} ResNet18 trained on SVHN (top) and CIFAR-10 (bottom) with \gls*{snl};
    \textbf{2nd col:} ResNet18 trained on UTKFace age (top) and race (bottom) labels with \gls*{snl}; 
    \textbf{3rd col:} ResNet34 trained on UTKFace age (top) and race (bottom) labels with \gls*{snl};
    \textbf{4th col:} ResNet18 trained on UTKFace age (top) and race (bottom) labels with \gls*{dr};
    The results show that performance on minority groups is affected disparately by \gls*{relu} linearization across choices of datasets, model architectures, and linearization methods.}
    \label{fig:lineplots_snl_datasets_archs}
\end{figure*}
\textbf{Linearization Methods.}
The paper considers two ReLU linearization techniques, representing the current state of the art: 
\glsreset{snl}\gls*{snl} \cite{ChoJRGH2022ICML} and \glsreset{dr}\gls*{dr} \cite{JhaGGR2021ICML}. 
Both methods aim to linearize \gls*{relu}s in neural networks, but their approaches differ: \gls*{snl} selectively linearizes specific \gls*{relu}s, whereas \gls*{dr} provides for a much coarser linearization strategy. In particular, \gls*{dr} does not allow to control the ``distribution'' of the \gls*{relu} functions across the architecture, but only to linearize blocks of contiguous nodes. This fundamental difference is expected to result in distinct patterns of disparity observed with each method. As a matter of fact, \gls*{dr} is not inherently data-driven, and its performance is not only affected by the number of residual \gls*{relu} functions, but also by the block that the practitioner decides to linearize. In turn, this results in different linearized model configurations as reported in \cref{fig:dr_glob_acc_utk}\footnote{We also attempted to incorporate results from SENet \cite{KunduLZLB2023ICLR}, but faced challenges. Although a validation code for the published models is available, a public implementation of the linearization algorithm is missing. Despite our efforts, we could not obtain the code or reproduce the results. Given that SENet aims to preserve global accuracy, we suspect it may encounter fairness issues similar to other frameworks considered.}.
 
In the following, the term \lq\lq \gls*{relu} budget" is used to refer to the number of \gls*{relu}s retained in a \gls*{relu}-linearized model across both methods. However, note that specifying a desired ReLU budget is feasible only with \gls*{snl}. In contrast, with \gls*{dr}, the number of retained ReLUs depends on the specific layers linearized. Additionally, while \gls*{relu} linearization may lead to a loss in utility, notice that the models and \gls*{relu} budgets used for the analysis yield a minor decrease in global test accuracies (as shown in Figure \ref{fig:GlobalTestAccs_UTKFace_age_SNL_RN34}) indicating that these configurations represent practical, deployable ReLU-linearized models.

\subsection{Impact on Accuracy}
\label{sec:impact_on_accuracy}
In the primary empirical findings, the study highlights the relative decreases in test accuracy for various groups under different ReLU budgets, as shown in \cref{fig:lineplots_snl_datasets_archs}. For datasets with more classes like CIFAR-10 and SVHN, the analysis focuses on the two most and two least represented classes in the test sets to maintain clarity and prevent plot overcrowding. The \emph{relative drop in test accuracy} serves as a measure of accuracy loss due to ReLU linearization. The larger this value, the greater the performance decline. Let $\bar{J}(\bm{\theta}; S) = \nicefrac{1}{|S|} \sum_{(\bm{x}, y)\in S} \indicatorfunc{f_{\bm{\theta}}(\bm{x}) = y}$, measure the classifier $\bm{\theta}$ accuracy over set $S$, then the relative drop in test accuracy for group $a$ is measured as: 
\[
    \frac{\bar{J}(\bm{\theta}; S^a) - \bar{J}(\tilde{\bm{\theta}}_r; S^a)}{\bar{J}(\bm{\theta}; S^a)} \times 100.
\]
The results reveal a consistent trend: majority groups generally show resilience to ReLU linearization while minority groups experience increasingly significant accuracy losses with more linearization. This pattern holds true across various architectures and datasets. For example, using SNL on ResNet-34 trained on UTKFace with race labels, white individuals experience almost no impact in accuracy drop across different ReLU linearization budgets, while the ``other'' race group suffers an almost 5-fold accuracy drop. Similarly, for \gls*{dr} on ResNet-18 trained on the same dataset, the accuracy for white individuals remains almost unaffected while the ``other'' race group suffers an almost 10-fold accuracy drop, with a near $100\%$ accuracy drop for the lowest \gls*{relu} budget. Indeed, a similar observation holds for \gls*{dr} for UTKFace with age labels.

\begin{figure*}[htb!]
    \centering
    \begin{subfigure}{0.24\textwidth}
        \includegraphics[width=\linewidth]{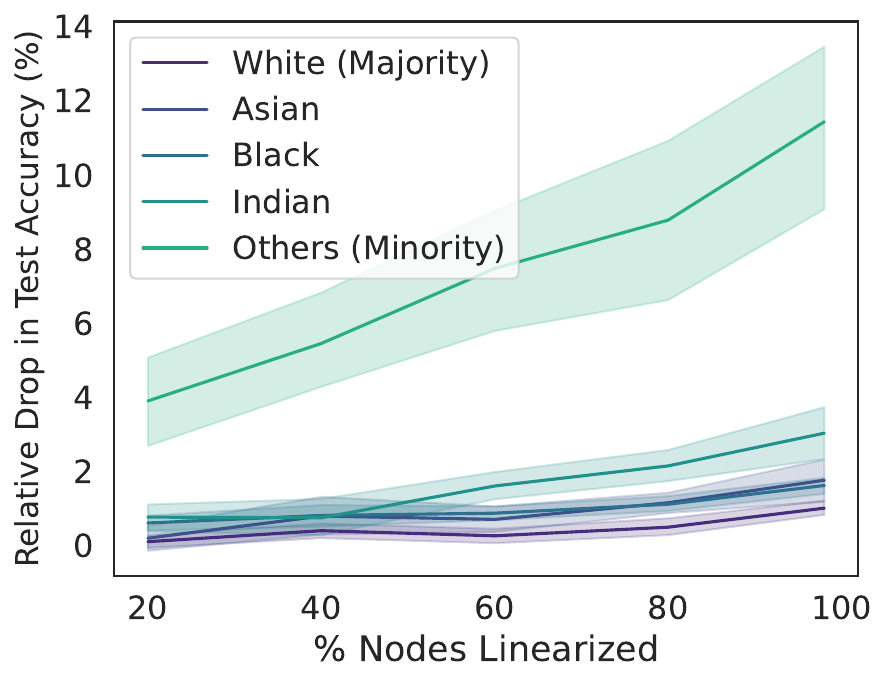}
        \caption{SNL on UTKFace (race) for VGG-19.}
        \label{fig:vgg_utk_race_snl}
    \end{subfigure}
    \begin{subfigure}{0.24\textwidth}
        \centering
        \includegraphics[width=\linewidth]{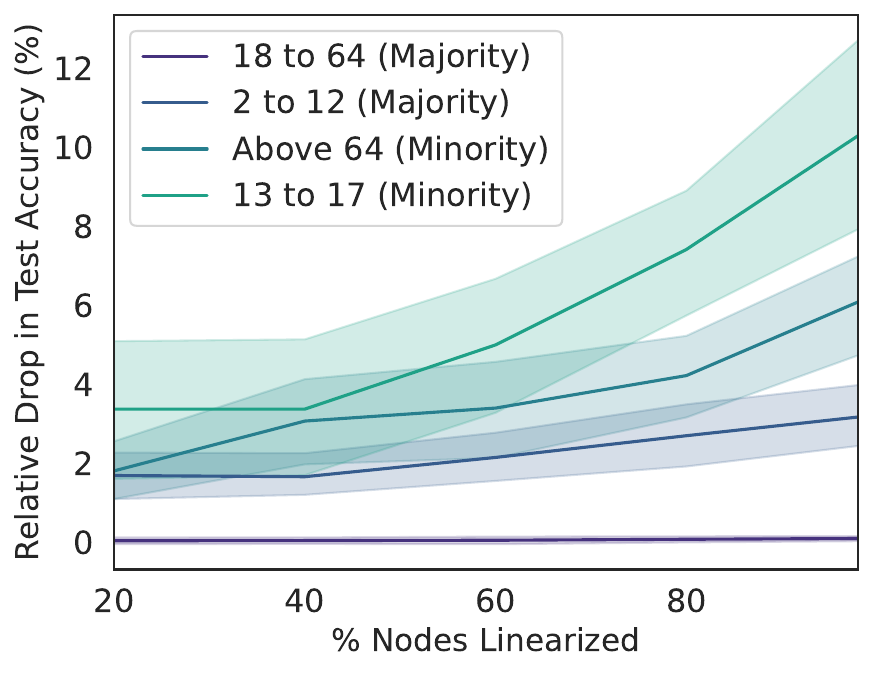}
        \caption{SNL on UTKFace (age) for VGG-19.}
        \label{fig:vgg_utk_age_snl}
    \end{subfigure}
    \begin{subfigure}{0.24\textwidth}
        \includegraphics[width=\linewidth]{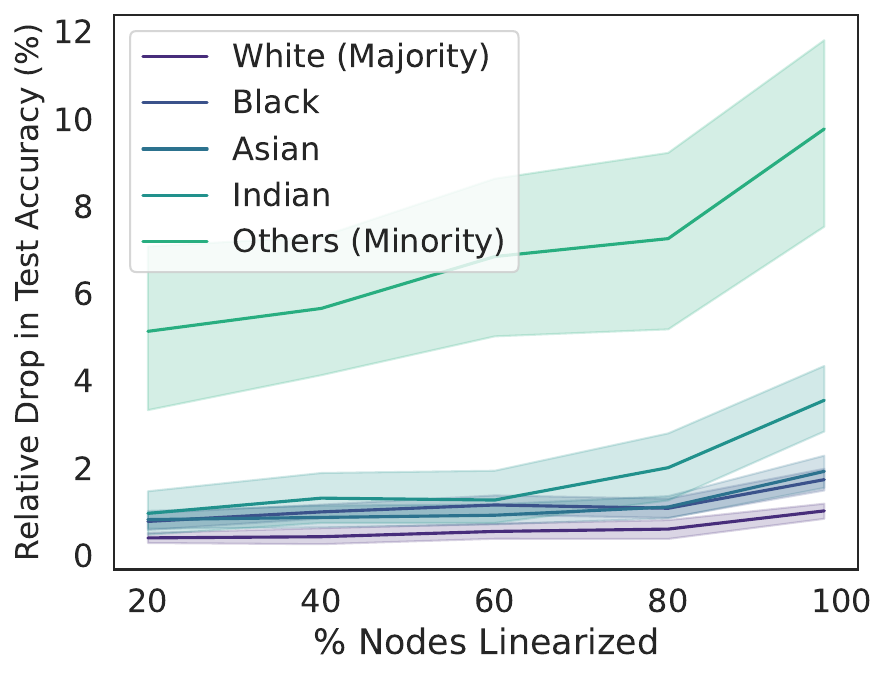} 
        \caption{SNL on UTKFace (race) for WideResNet.}
        \label{fig:wideresnet_utk_race_snl}
    \end{subfigure}
    \begin{subfigure}{0.24\textwidth}
        \centering
        \includegraphics[width=\linewidth]{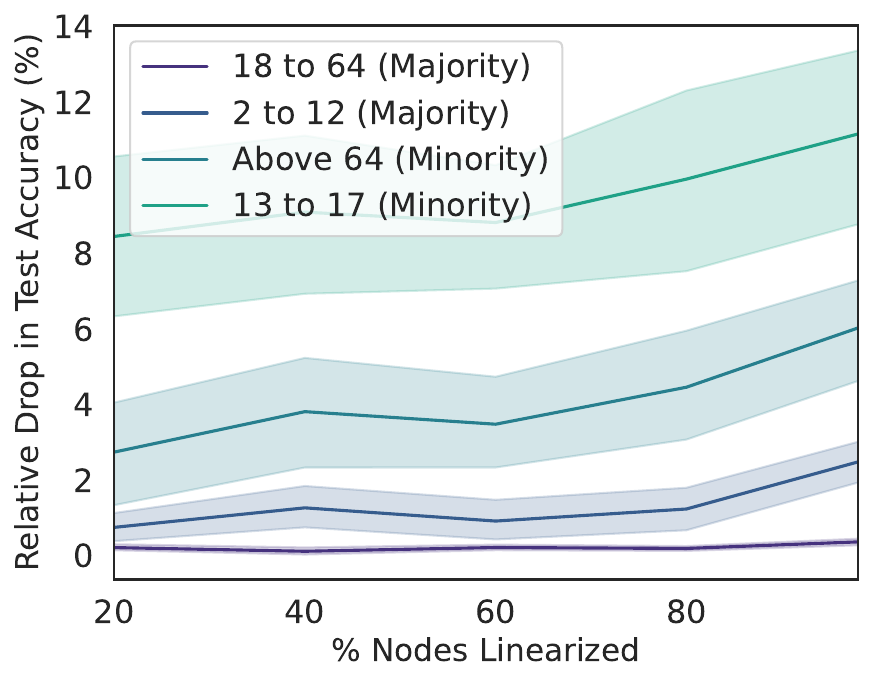}
        \caption{SNL on UTKFace (age) for WideResNet.}
        \label{fig:wideresnet_utk_age_snl}
    \end{subfigure}
\caption{
Results for VGG-19 and WideResNet models on UTKFace with \gls*{snl}.
We observe that the performance is consistent with the results obtained for the ResNet18 and
ResNet34 models: more linearization produces higher accuracy loss for underrepresented groups, while the accuracy remains almost unchanged in the majority group.
We observe that the phenomenon of disparity in the accuracy for underrepresented groups is
evident on WideResNet-22-8, a wider variant of ResNet.
}
\label{fig:vgg_wideresnet_utkface}
\end{figure*}

Recall that Theorem \ref{prop:bound_residual_loss} highlighted the connection between a drop in loss (as a differentiable proxy for accuracy) experienced by a given group, under a certain ReLU linearization level, and two key characteristics: 
the gradient norms associated with such groups at convergence and the distance to the decision boundary. While these results hold for a restricted classes of functions, the next two sections provide further evidence on such relationships on highly non linear models for a variety of linearization techniques and architectures. 
Further experiments on different architectures and wider variants of the ResNet models are 
reported in \cref{fig:vgg_wideresnet_utkface}.

\subsection{Gradient Norms and Fairness}
\label{sec:GradNorm}
\begin{figure*}[!h]
    \centering
    \hfill
    \includegraphics[width=0.30\linewidth]{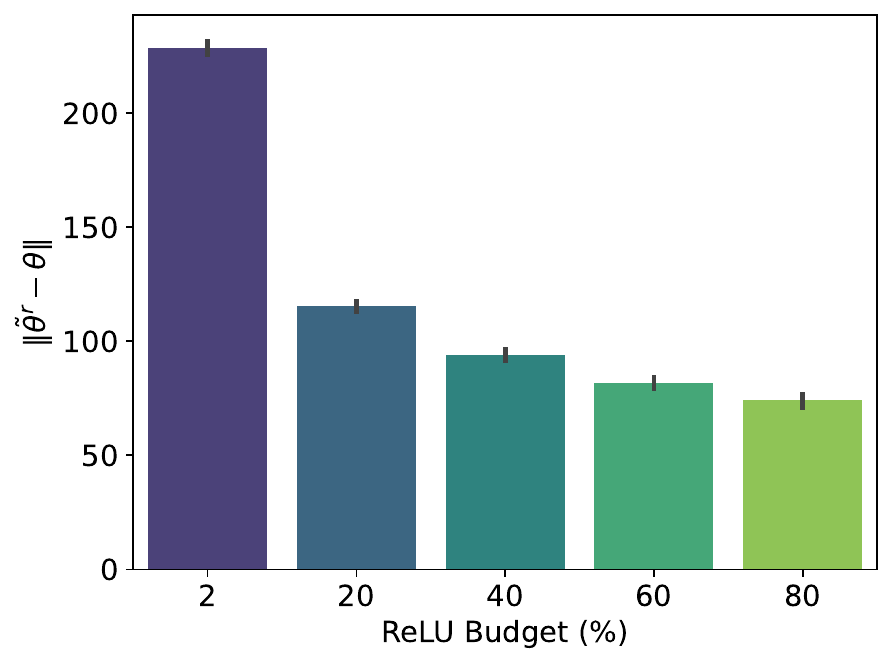}
    \hfill
    \includegraphics[width=0.30\linewidth]{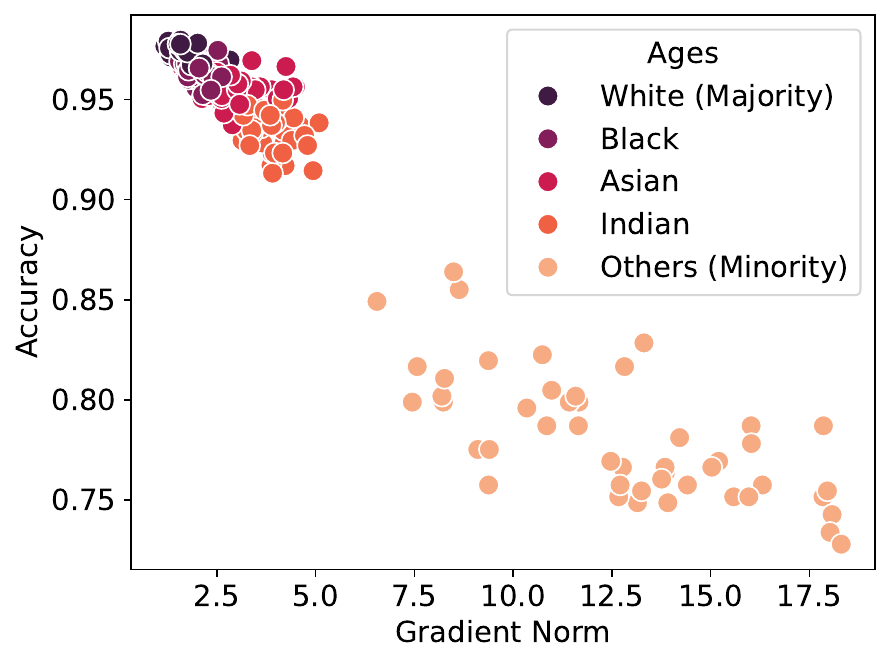}
    \hfill
    \includegraphics[width=0.30\linewidth]{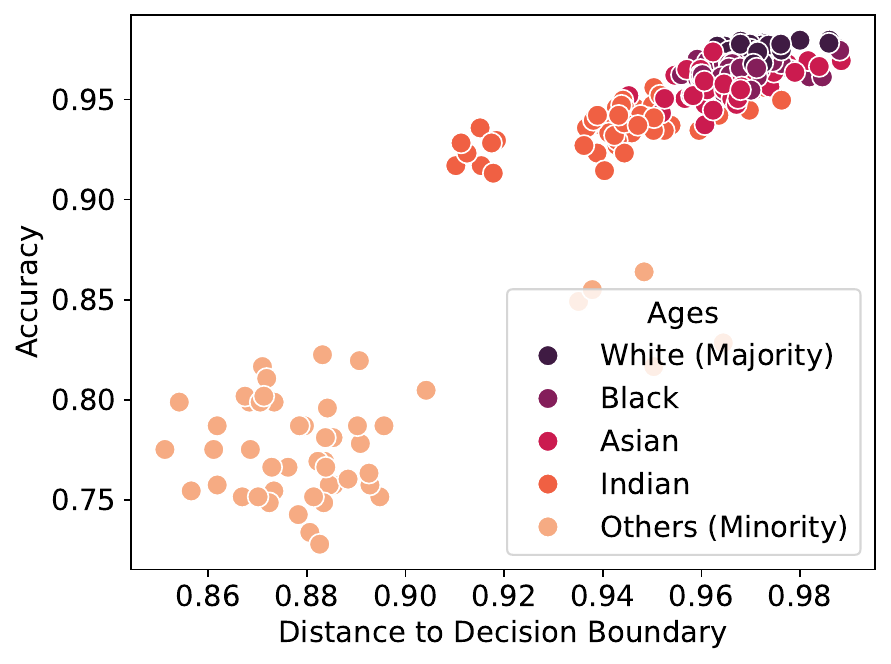}
    \hfill
    \vspace{-6pt}
    \caption{\gls*{snl} on UTKFace with race labels and ResNet-18: \textbf{left:} $\Vert\paramslin^{r}-\paramsrelu\Vert$ vs. \gls*{relu} budget; \textbf{middle:} group accuracy vs.~gradient norm for 2\% \gls*{relu}s retained; \textbf{right:} group accuracy vs.~distance to decision boundary for 2\% \gls*{relu}s retained.}
    \label{fig:TestAccsandOtherMetrics}
\end{figure*}

Firstly, we notice Theorem \ref{prop:bound_residual_loss} 
highlights how both the group gradient and the group Hessian terms are multiplicatively influenced by the distance $\|\tilde{\bm{\theta}} - \bm{\theta}\|$ between the model with reduced ReLU functions and the original model. 
\cref{fig:TestAccsandOtherMetrics} (left) illustrates that this distance increases with the increase in linearized ReLU functions, suggesting that the reduction in ReLU activations amplifies the effect on the gradient norm. Next, \cref{fig:TestAccsandOtherMetrics} (middle) presents a scatterplot that contrasts accuracy against gradient norms at the highest level of linearization examined. It reveals an inverse relationship between gradient norms and accuracy: data points for the majority group align with lower gradient norms and higher accuracy, whereas points for minority groups are linked with higher gradient norms and lower accuracy.
Taken together, the increasing distance between reduced and original model parameters, alongside the varied impact on gradient norms across different groups, highlight the dynamics of gradient norms as a factor explaining the exacerbation of unfairness post-ReLU linearization.

\subsection{Distance to the Decision Boundary and Fairness}
\label{sec:DecisionBoundary}
Next, we look into the influence of the group Hessian on the accuracy reduction resulting from ReLU linearization. Recall that Proposition \ref{prop:max_eig} establishes a link between the group Hessian and the distance to the decision boundary: notably, a smaller distance (or more proximity) to the decision boundary correlates with higher bounds for the associated group Hessian. This relationship renders the distance to the decision boundary an insightful, interpretable metric. 
\cref{fig:TestAccsandOtherMetrics} (right) elucidates this concept. It illustrates the relationship between each test sample's distance to the decision boundary and its accuracy. The data points are color-coded according to group membership, revealing a strong correlation between the distance to the boundary and accuracy levels. Specifically, groups identified as majorities tend to have a larger distance to the boundary (and high accuracy), whereas minority groups exhibit a smaller distance.

This observation, in conjunction with what observed in \cref{fig:TestAccsandOtherMetrics} (left), underscores the key relation between distance to the decision boundary plays and fairness issues following ReLU linearization. 
Understanding of this interplay is crucial to design the proposed mitigation technique, reviewed in the next section.

\section{Mitigation and Impact on Fairness}
\label{sec:mitigation}

So far, we have presented an in-depth analysis of the disparate effects that linearization techniques produce on the fairness of unbalanced datasets, showing how reduced \gls*{relu} budgets negatively affect the accuracy over underreprested groups. We will now present a solution to extend \gls*{relu} linearization to contexts where fairness matters along with \acrlong*{pi}.
Recall from \cref{prop:max_eig} the relationship between a group's Hessian eigenvalues, their proximity to the decision boundary, and the corresponding errors made by the model for that group. Since, as shown in the previous section, the distance to the decision boundary of a sample directly relates to their accuracy drop across various ReLU linearizaiton rates, 
our strategy leverages this intuition and introduces ``fairness regularizers'' to equalize the distances to the decision boundary of various groups.  These regularizers, capture the differences in losses between groups and population, (and by proxy the distance to the boundary). The method leverages Lagrangian duality principles \cite{FvHMTBL:ecml20} for implementation and it works off-the-shelf with the \gls*{kd} based finetuning step, offering a seamless integration into existing algorithms. 

\cref{alg:mitigation} reports the proposed fairness-aware finetuning method, replacing the standard finetuning step of \gls*{snl} and \gls*{dr}. Let us introduce the main notation.  Therein, $\losskd:\R^C\times\R^C\times\R^{d}\to\R$ is the loss function associated with the \gls*{kd} finetuning, which, in addition to a batch of data $B$, takes as input two classifiers $h_{\paramslin}^{r}$ and $h_{\paramsrelu}$, and returns a scalar; 
$\bm{\lambda_e}$ is the vector of Lagrangian multipliers, each associated to a fairness \emph{constraint violation}, expressed as the distance between the loss computed over a group and that computed over the whole dataset. These multipliers are updated at each epoch $e$;  
$\mu$ is the multiplier associated with the Lagrangian update step; and $\alpha$ is the learning rate. Furthermore, we call $\losskd_{a}$ the loss function, when evaluated over the samples of a specific group $a\in\aset$. 
Crucially, the proposed algorithm allows the student model, i.e., the linearized one, to be trained with knowledge flowing from the original model, while at the same time being constrained against excessive degradation of performance over underrepresented groups.

\begin{algorithm}[!t]
    \caption{Fairness Mitigation for ReLU Linearization}\label{alg:mitigation}
    \begin{algorithmic}
    \REQUIRE $h_{\paramslin^{r}}$, $h_{\paramsrelu}$, Multiplier $\mu>0$, step size $\lr$,\\ 
    loss function $\losskd$, group-wise loss function $\losskd_{a}$,\\
    dataset $\trainingsetshort$, set of groups $\aset$
    \STATE $\bm{\lambda}_1\gets (0)_{a\in \aset}$ \COMMENT{array of zeros with size $|\aset|$}
    \FOR{epoch $e=1,2,\cdots$}
    \FORALL{mini-batch $B\subseteq \trainingsetshort$}
    \STATE $\rho_{\paramslin^r,\paramsrelu,B}\gets \vert\losskd(h_{\paramslin^{r}},h_{\paramsrelu},B)-\losskd_{a}(h_{\paramslin^{r}},h_{\paramsrelu},B)\vert$
    \STATE $\paramslin^{r}\gets\paramslin^{r} - \alpha\nabla_{\paramslin^{r}}\left(\mathcal{L}(h_{\paramslin^{r}},h_{\paramsrelu},B) + \bm{\lambda}_{e}^\top\rho_{\paramslin^r,\paramsrelu,B}\right)$
    \ENDFOR
    \STATE ${\bm{\lambda}_{e+1}\gets\bm{\lambda}_e + \mu\vert\losskd(h_{\paramslin^{r}},h_{\paramsrelu},\trainingsetshort)-\losskd_{a}(h_{\paramslin^{r}},h_{\paramsrelu},\trainingsetshort)\vert}$
    \ENDFOR
    \end{algorithmic}
\end{algorithm}
Let $\bar{J}(\paramslin^{r}; S) = \nicefrac{1}{|S|} \sum_{(\bm{x}, y)\in S} \indicatorfunc{f_{\paramslin^{r}}(\bm{x}) = y}$, measure the classifier $\theta$ accuracy over set $S$, then the relative drop in test accuracy for group $a$ is measured as: 
\[
    \frac{\bar{J}(\paramslin^{r}; S^a) - \bar{J}(\paramslin^{r}_{mit}; S^a)}{\bar{J}(\paramslin^{r}; S^a)} \times 100,
\]
where, with a slight abuse of notation, $\paramslin^{r}$ represents the model parameter before the mitigation, and, $\paramslin^{r}_{mit}$ the parameter of the corresponding linearized model finetuned with our mitigation algorithm. 

\begin{figure*}[!htb]
    \centering
    \includegraphics[width=0.3\textwidth]{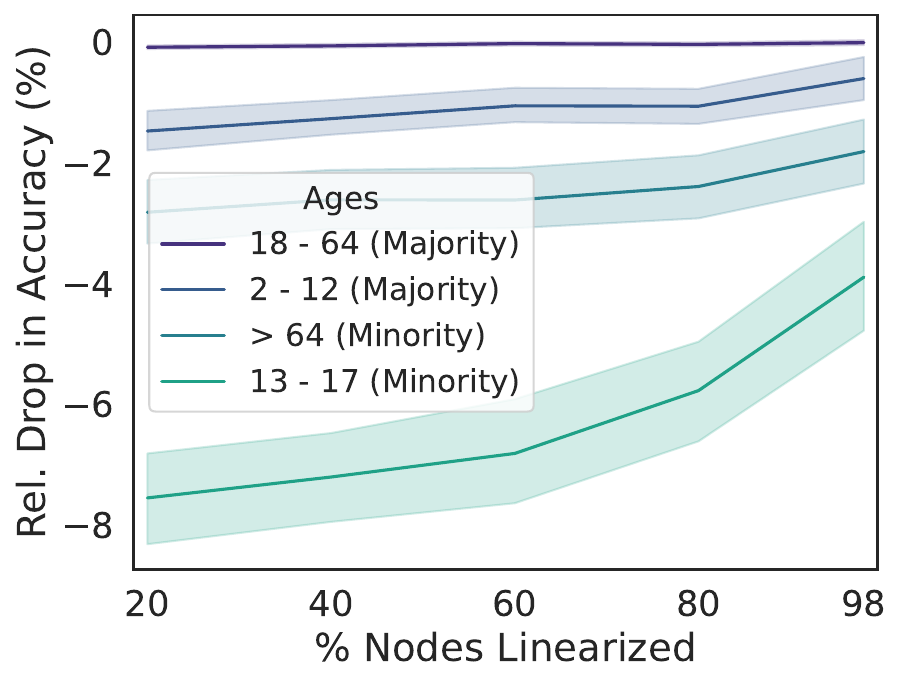} 
    \hspace{60pt}
    \includegraphics[width=0.3\textwidth]
    {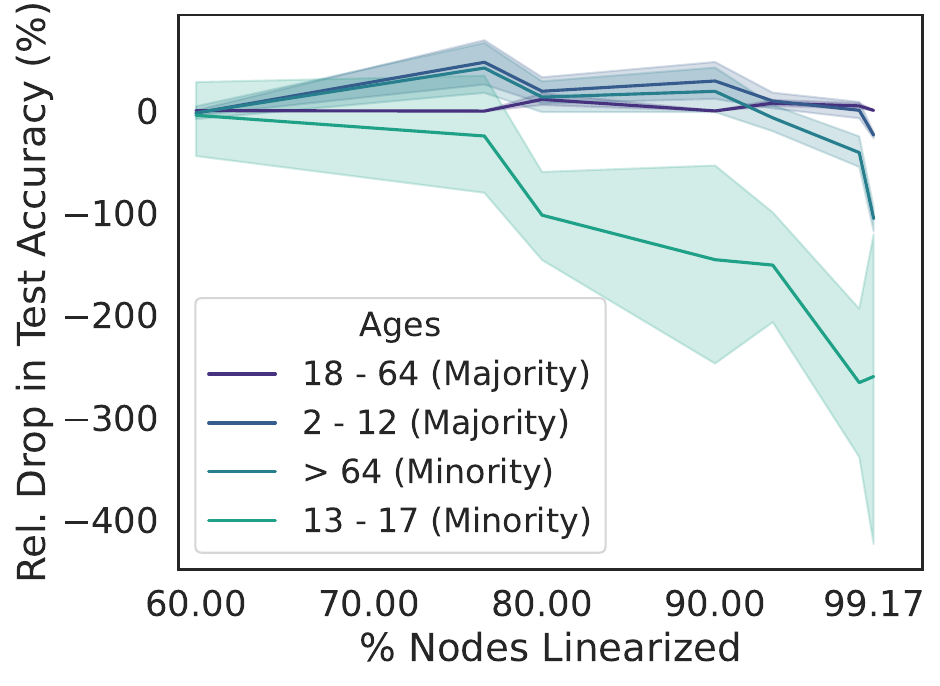}\\ 
    \includegraphics[width=0.3\textwidth]{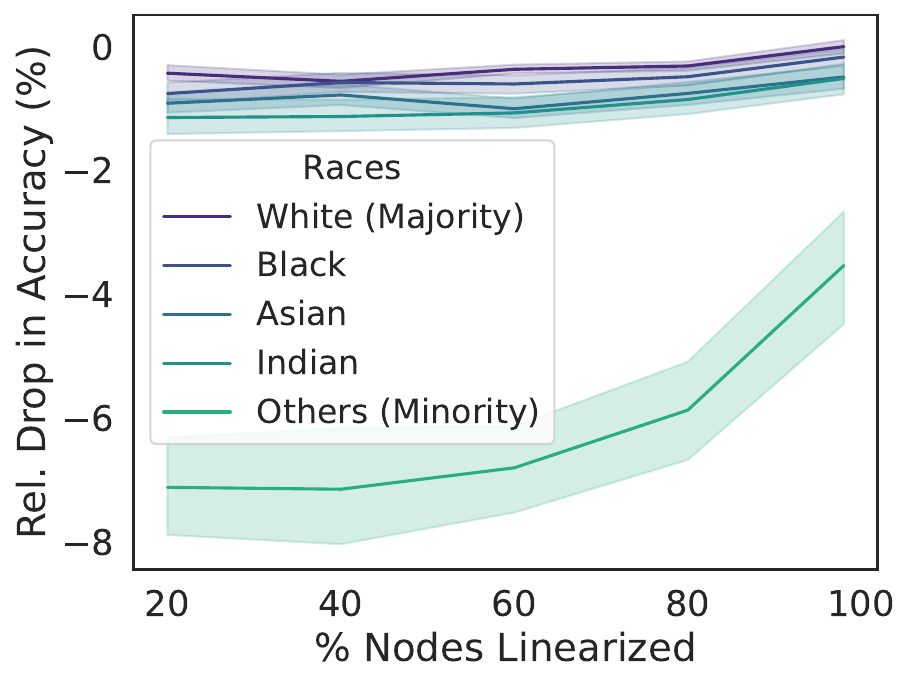}  
    \hspace{60pt}
    \includegraphics[width=0.3\textwidth]{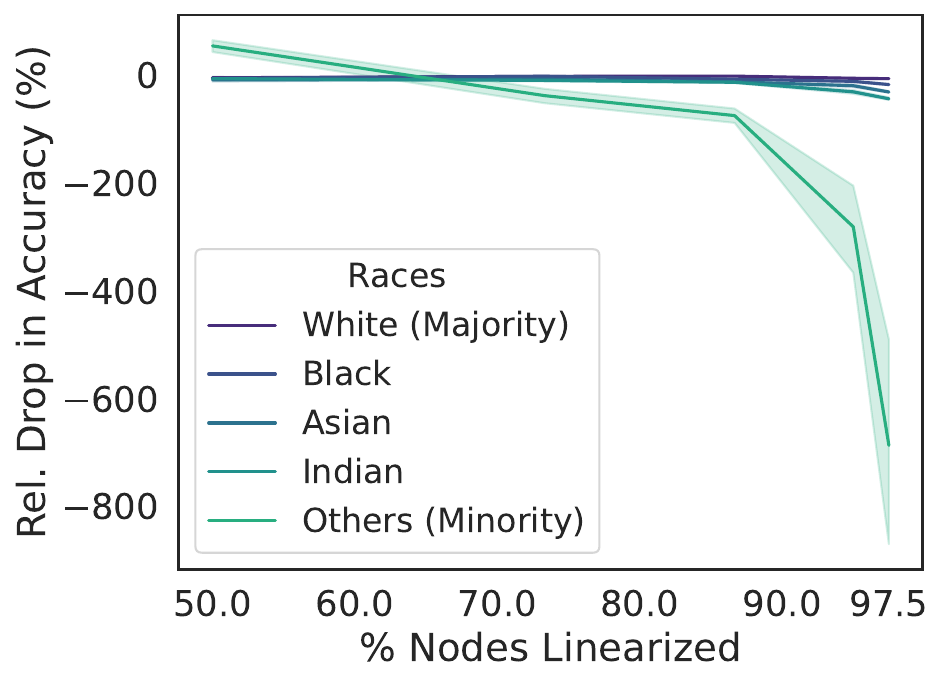}
    \vspace{-6pt}
    \caption{Mitigation: Relative test accuracy drop for different linearization methods. 
    \textbf{left:} ResNet18 trained on UTKFace age (top) and race (bottom) labels with \gls*{snl}; 
    \textbf{right:} ResNet18 trained on UTKFace age (top) and race (bottom) labels with \gls*{dr}. 
    }
    \label{fig:lineplots_accs_datasets_archs_mitigate}
\end{figure*}

\cref{fig:lineplots_accs_datasets_archs_mitigate} illustrates the effect of the mitigation strategy on the UTKFace dataset for both SNL (left) and \gls*{dr} (right) methods, with results segmented by age (top) and race (bottom) labels. 
Negative values indicate accuracy improvements for a group compared to the baseline accuracy of the original model at the same level of ReLU reduction. 
There are two key aspects: First, the mitigation consistently enhances accuracy across all groups and linearization levels tested, with minimal negative effects on majority classes.
Second, notice how the minority groups improve their relative performance, a trend that is especially evident for \gls*{dr} at higher levels of ReLU linearization. Furthermore, there is an observable reduction in gradient norms and an augmentation in the distance to the decision boundary for minority groups under the mitigation approach (see  \cref{sec:app_mitigation_gradnorms}). 
These results illustrate the potential for effective mitigations to significantly improve the performance for minority groups without detriment to majority groups, thus substantially reducing the unfairness produced by ReLU linearization.

Finally, an ablation study on the impact of the 
parameter $\mu$ in the mitigation algorithm is 
presented in \cref{sec:app_mitigation_mu_ablation}.

\section{Discussion}

\Cref{fig:lineplots_accs_datasets_archs_mitigate} demonstrates the effectiveness of the mitigation method, highlighting the importance of choosing the right approach for practitioners aiming to enforce fairness predictably. A data-driven ReLU linearization method like SNL, which carefully considers the impact of linearization on model utility and fairness, stands out as preferable over simpler methods such as \gls*{dr}. In \gls*{dr}, ReLUs may be linearized arbitrarily across any layer without regard for potential impacts on the model performance.

This is evident in \cref{fig:lineplots_snl_datasets_archs}, where the smallest groups under \gls*{dr} exhibit significantly poorer performance compared to others and when SNL is applied. The unpredictable utility and fairness outcomes with \gls*{dr} make it a less suitable option for achieving fair models predictably. In contrast, SNL offers consistent and predictable outcomes, enabling practitioners to more easily apply fair ReLU linearization and select suitable parameters for effective mitigation.

\textbf{Limitations.}
Our theoretical framework proposes a bound for the residual loss 
of a DNN model and its linearized counterpart, elucidating their 
relationship with group norms, group Hessians, and the norm of 
the weight distance between the two models' parameters. 
While our analysis introduces proxies to articulate the connection 
between group size, linearized ReLU proportion, and accuracy drop, 
examining the actual relationship between these quantities remains 
an open question. Addressing this gap presents an avenue for future 
research, with an opportunity to formalize fundamental limits 
of disparate impact mitigation strategies through non-trivial lower
bounds on loss differences.
We note that the proposed mitigation strategy necessitates 
access to protected group information. Although this information 
is frequently available, the requirement may restrict the 
applicability of the proposed mitigation in contexts where such 
information may be missing. To address this limitation, future 
research could explore the integration of our mitigation approach 
with tailored loss functions to promote fairness without 
relying on demographic information (cf. \citet{LahotiBCLPTWC2020NeurIPS}) or with a context 
of distributionally robust objectives (cf. \citet{HashimotoSNL2018ICML}).


\section{Conclusion}
\label{sec:conclusion}
This study 
observed that 
while ReLU linearization strategies effectively reduce computational costs and inference times, they inadvertently exacerbate accuracy disparities across different subgroups, particularly affecting underrepresented ones. This phenomenon was found consistent across various datasets and  architectures. 
The empirical analysis was grounded in theoretical insights, which highlighted two key factors responsible for the observed unfairness. We showed that as \gls*{relu} functions are approximated, the gradient for underrepresented groups tend to increase, suggesting that the model becomes more sensitive to input variations for these groups, leading to a higher misclassification rates. This effect is exacerbated for underrepresented groups due to their typically smaller distance from the decision boundary in the space that these models operate within. This distance to the boundary, when combined with altered gradient norms due to \gls*{relu} linearization, results in a disproportionate impact on the fairness of the model's outcomes. 

Motivated by such observations, the paper proposed a mitigation solution which acts on regularizing the finetuning step of ReLU linearization strategies through a Lagrangian dual approach. This simple solution was found effective in balancing the computational benefit of ReLU linearization with the imperative of fairness. 

\newpage
\section*{Acknowledgements}
This research is partially supported by NSF grants 2133169, 2232054 and NSF CAREER Award 2143706. Fioretto is also supported by an Amazon Research Award and a Google Research Scholar Award. Its views and conclusions are those of the authors only.
\section*{Impact Statement}
Our contributions shed light on the Matthew effect within accuracy reduction caused by \gls*{relu} linearization, offering a mathematical intuition of the observed disparities, demonstrating the widespread presence of these effects across algorithms and models, and proposing a viable mitigation strategy. The proposed strategy not only addresses fairness concerns but also manages to preserve overall accuracy to a commendable extent.

The evidence presented, particularly the stark accuracy disparities illustrated in \autoref{fig:accuracy_assesment_intro}, calls for a reevaluation of model optimization strategies, especially in the use of datasets reflecting real-world diversity. The findings emphasize the necessity for approaches that consider the differential impact of linearization on various subgroups, advocating for a more inclusive and fair machine learning ecosystem.

\bibliographystyle{bst}
\bibliography{references}

\newpage
\appendix
\onecolumn

\section{Additional Theoretical Results}
\subsection{Underrepresented groups have larger gradients.}
\label{subsec:underrepresented_groups_have_larger_gradients}

Let us assume $a,b\in\aset$ such that with $\left|S^a\right| \geq\left|S^b\right|$. Then $\left\|\boldsymbol{g}_a^{\ell}\right\| \leq\left\|\boldsymbol{g}_b^{\ell}\right\|$.

\begin{proof}
    Let us assume convergence of the model with parameters $\boldsymbol{\phi}$ to a (local) minimum. Then, it holds that:
    $$
    \begin{aligned}
    \nabla \mathcal{L}(\boldsymbol{\phi} ; S) & =\sum_{a \in \mathcal{A}} \frac{\left|S^a\right|}{|S|} \nabla J\left(\boldsymbol{\phi} ; S^a\right) \\
    & =\frac{\left|S^a\right|}{|D|} \boldsymbol{g}_a^{\ell}+\frac{\left|S^b\right|}{|D|} \boldsymbol{g}_b^{\ell}=\mathbf{0}
    \end{aligned}
    $$

    Thus, $\boldsymbol{g}_a^{\ell}=-\frac{\left|S^b\right|}{\left|S^a\right|} \boldsymbol{g}_b$. Hence $\left\|\boldsymbol{g}_a^{\ell}\right\|=\frac{\left|S^b\right|}{\left|S^a\right|}\left\|\boldsymbol{g}_b^{\ell}\right\| \leq\left\|\boldsymbol{g}_b^{\ell}\right\|$, because $\left|S^a\right| \geq\left|S^b\right|$.

\end{proof}

\subsection{An upper-bound for the residual loss.}
\label{app:proof_1}
Theorem 1. The excessive loss of a group $a \in \mathcal{A}$ is upper bounded by
$$
R(a) \leq\left\|\boldsymbol{g}_a^{\ell}\right\| \times\|\paramslin-\paramsrelu\|+\frac{1}{2} \lambda\left(\boldsymbol{H}_a^{\ell}\right) \times\|\paramslin-\hat{\boldsymbol{\theta}}\|^2+O\left(\|\paramslin-\hat{\boldsymbol{\theta}}\|^3\right),
$$
where $\boldsymbol{g}_a^{\ell}=\nabla J\left(\paramsrelu ; S^a\right)$ is the vector of gradients associated with the loss function $\ell$ evaluated at $\paramsrelu$ and computed using group data $S^a, \boldsymbol{H}_a^{\ell}=\nabla^2 J\left(\paramsrelu ; S^a\right)$ is the Hessian matrix of the loss function $\ell$, at the optimal parameters vector $\paramsrelu$, computed using the group data $S^a$ (henceforth simply referred to as group hessian), and $\lambda(\Sigma)$ is the maximum eigenvalue of a matrix $\Sigma$.

Proof. Using a second order Taylor expansion around $\paramsrelu$, the excessive loss $R(a)$ for a group $a \in \mathcal{A}$ can be stated as:
$$
\begin{aligned}
R(a) & =J\left(\paramslin ; S^a\right)-J\left(\paramsrelu ; S^a\right) \\
& =\left[J\left(\paramsrelu ; S^a\right)+(\paramslin-\paramsrelu)^{\top} \nabla J\left(\paramsrelu ; S^a\right)+\frac{1}{2}(\paramslin-\paramsrelu)^{\top} \boldsymbol{H}_a^{\ell}(\paramslin-\paramsrelu)+O\left(\|\paramsrelu-\paramslin\|^3\right)\right]-J\left(\paramsrelu ; S^a\right) \\
& =(\paramslin-\paramsrelu)^{\top} \boldsymbol{g}_a^{\ell}+\frac{1}{2}(\paramslin-\paramsrelu)^{\top} \boldsymbol{H}_a^{\ell}(\paramslin-\paramsrelu)+O\left(\|\paramsrelu-\paramslin\|^3\right)
\end{aligned}
$$

The above, follows from the loss $\ell(\cdot)$ being at least twice differentiable, by assumption.
By Cauchy-Schwarz inequality, it follows that
$$
(\paramslin-\paramsrelu)^{\top} \boldsymbol{g}_a^{\ell} \leq\|\paramslin-\paramsrelu\| \times\left\|\boldsymbol{g}_a^{\ell}\right\| .
$$

In addition, due to the property of Rayleigh quotient we have:
$$
\frac{1}{2}(\paramslin-\paramsrelu)^{\top} \boldsymbol{H}_a^{\ell}(\paramslin-\paramsrelu) \leq \frac{1}{2} \lambda\left(\boldsymbol{H}_a^{\ell}\right) \times\|\paramslin-\paramsrelu\|^2
$$




\subsection{An upper-bound for the maximum eigenvalue of the group Hessian}
\label{app:proof_2}

Let $f_{\paramsrelu}$ be a binary classifier trained using a binary cross entropy loss. For any group $a \in \mathcal{A}$, the maximum eigenvalue of the group Hessian $\lambda\left(\boldsymbol{H}_a^{\ell}\right)$ can be upper bounded by:
$$
\lambda\left(\boldsymbol{H}_a^{\ell}\right) \leq \frac{1}{\left|S^a\right|} \sum_{(\boldsymbol{x}, y) \in S^a} \underbrace{\left(h_{\paramsrelu}(\boldsymbol{x})\right)\left(1-h_{\paramsrelu}(\boldsymbol{x})\right)}_{\text {Proximity to decision boundary }} \times\left\|\nabla h_{\paramsrelu}(\boldsymbol{x})\right\|^2+\underbrace{\left|f_{\paramsrelu}(\boldsymbol{x})-y\right|}_{\text {Accuracy }} \times \lambda\left(\nabla^2 h_{\paramsrelu}(\boldsymbol{x})\right) .
$$

\begin{proof} First notice that an upper bound for the Hessian loss computed on a group $a \in \mathcal{A}$ can be derived as:
\begin{align}
\label{eq:hessian_first_upperbound}
\lambda\left(\boldsymbol{H}_a^{\ell}\right)=\lambda\left(\frac{1}{\left|S^a\right|} \sum_{(\boldsymbol{x}, y) \in S^a} \boldsymbol{H}_x^{\ell}\right) \leq \frac{1}{\left|S^a\right|} \sum_{(\boldsymbol{x}, y) \in S^a} \lambda\left(\boldsymbol{H}_x^{\ell}\right)
\end{align}
where $\boldsymbol{H}_{\boldsymbol{x}}^{\ell}$ represents the Hessian loss associated with a sample $\boldsymbol{x} \in S^a$ from group $a$. The above follows Weily's inequality which states that for any two symmetric matrices $A$ and $B, \lambda(A+B) \leq$ $\lambda(A)+\lambda(B)$.
Next, we will derive an upper bound on the Hessian loss associated to a sample $\boldsymbol{x}$. First, based on the chain rule a closed form expression for the Hessian loss associated to a sample $\boldsymbol{x}$ can be written as follows:
$$
\boldsymbol{H}_{\boldsymbol{x}}^{\ell}=\nabla^2 \ell\left(f_{\paramsrelu}(\boldsymbol{x}), y\right)\left[\nabla f_{\paramsrelu}(\boldsymbol{x})\left(\nabla f_{\paramsrelu}(\boldsymbol{x})\right)^{\top}\right]+\nabla \ell\left(f_{\paramsrelu}(\boldsymbol{x}), y\right) \nabla^2 f_{\paramsrelu}(\boldsymbol{x}) .
$$

The next follows from that
$$
\begin{aligned}
& \nabla \ell\left(f_{\paramsrelu}(\boldsymbol{x}), y\right)=\left(f_{\paramsrelu}(\boldsymbol{x})-y\right), \\
& \nabla^2 \ell\left(f_{\paramsrelu}(\boldsymbol{x}), y\right)=f_{\paramsrelu}(\boldsymbol{x})\left(1-f_{\paramsrelu}(\boldsymbol{x})\right) .
\end{aligned}
$$

Applying the Weily inequality again on the R.H.S. of Equation 12, we obtain:

\begin{align}
\label{eq:Weily_hessian}
\nonumber\lambda\left(\boldsymbol{H}_{\boldsymbol{x}}^{\ell}\right) & \leq f_{\paramsrelu}(\boldsymbol{x})\left(1-f_{\paramsrelu}(\boldsymbol{x})\right) \times\left\|\nabla f_{\paramsrelu}(\boldsymbol{x})\right\|^2+\lambda\left(f_{\paramsrelu}(\boldsymbol{x})-y\right) \times \nabla^2 f_{\paramsrelu}(\boldsymbol{x}) \\
& \leq f_{\paramsrelu}(\boldsymbol{x})\left(1-f_{\paramsrelu}(\boldsymbol{x})\right) \times\left\|\nabla f_{\paramsrelu}(\boldsymbol{x})\right\|^2+\left|f_{\paramsrelu}(\boldsymbol{x})-y\right| \lambda\left(\nabla^2 f_{\paramsrelu}(\boldsymbol{x})\right)
\end{align}

The statement of \cref{prop:max_eig} is obtained combining \cref{eq:hessian_first_upperbound} with \cref{eq:Weily_hessian}.



\end{proof}

\section{Examples}
\subsection{On linearized models and the upper-bound in \cref{prop:relu_upper_bound}: an empirical standpoint.}
\label{sec:example-prop4.2}

\begin{figure}[!htb]
         \centering
         \begin{subfigure}[b]{.35\columnwidth}
             \centering
             \includegraphics[width=\textwidth]{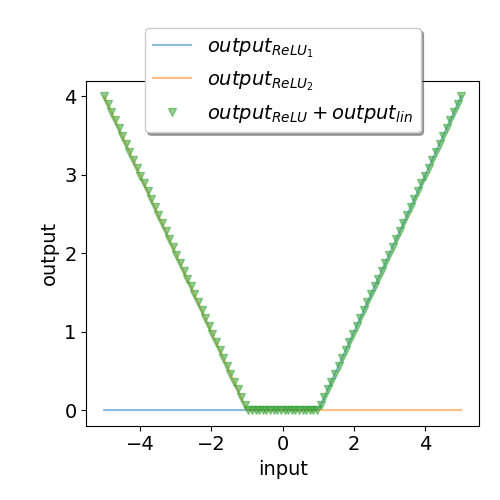}
             \caption{Model with $2$ \gls*{relu} nodes.\\
             \phantom{phantom}}
             \label{fig:rc1}
        \end{subfigure}
        \vspace{24pt}
        \begin{subfigure}[b]{.35\columnwidth}
             \centering
             \includegraphics[width=\textwidth]{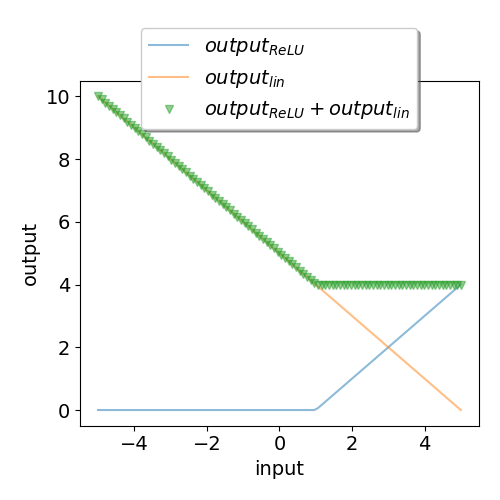}
             \caption{Model with $1$ \gls*{relu} node and a linearized one.}
             \label{fig:rc2}
         \end{subfigure}
        \vskip -0.2in
        \caption{The \gls*{relu} network in \cref{fig:rc1} can attain at most $3$ linear pieces, while the linearized network in \cref{fig:rc2} can attain at most $2$ linear pieces.}
        \label{fig:rc}
\end{figure}

In general, linearized models may be far away from reaching the bound in \cref{prop:relu_upper_bound}.
It is easy to see how this bound cannot be reached in many cases when one or more \gls*{relu} activation are replaced with linear activation functions. For instance let us consider a \gls*{relu} network with a single hidden layer and width $\omega_1=2$. 
According to \cref{prop:relu_upper_bound}, this network can attain at most $3$ linear pieces. 
Indeed, let us consider bounded inputs $\x\in[-b, b]$ such that $b\in\R^{+}$. By properly optimizing weights and biases, the two \gls*{relu} nodes can create the orange and blue linear pieces in \cref{fig:rc1}, that, when linearly combined in the output of the model (cf. green line in \cref{fig:rc1}), can attain at most $3$ linear pieces. On the contrary, if one of the two \gls*{relu} nodes is replaced with a linear node, the linear combination of the two nodes which is forwarded to the output, by definition, does not create any new breakpoint, only maintaining the two linear pieces dictated by the presence of the \gls*{relu} node. Therefore, the number of linear pieces is reduced to 2, as shown in \cref{fig:rc2}.

\section{Additional Empirical Results} 
\subsection{Settings}
Each of the results in this paper was produced using an A100 GPUs with 80 GB of GPU memory, up to 100 GB of RAM, and up to 10 Intel(R) Xeon(R) E5-2630 v3 CPUs each clocked at 2.40GHz.

For \gls*{snl}, we train each base model for 160 epochs, and then perform the \gls*{snl} finetuning step after \gls*{relu} linearization. These values are present in the official implementation of this algorithm provided by the authors of \cite{ChoJRGH2022ICML}, and we use this implementation off-the-shelf.

For \acrlong*{dr}, we use the official implementation provided by the authors of \cite{JhaGGR2021ICML} off-the-shelf which trains each model for 200 epochs.

\subsection{Accuracy drop analysis}

For SVHN and CIFAR-10, the classes corresponding to these labels are evident. The class correspondences for UTKFace with age and race labels are as follows.
\begin{itemize}
    \item \textbf{UTKFace with age labels:} 2-12, 13-17, 18-64, and over 64 years of age correspond to 1, 2, 3, and 4, respectively.
    \item \textbf{UTKFace with race labels:} White, black, Asian, Indian, and other races correspond to 0, 1, 2, 3, and 4, respectively.
\end{itemize}

\begin{figure}[!h]
    \centering
    \includegraphics[width=0.24\textwidth]{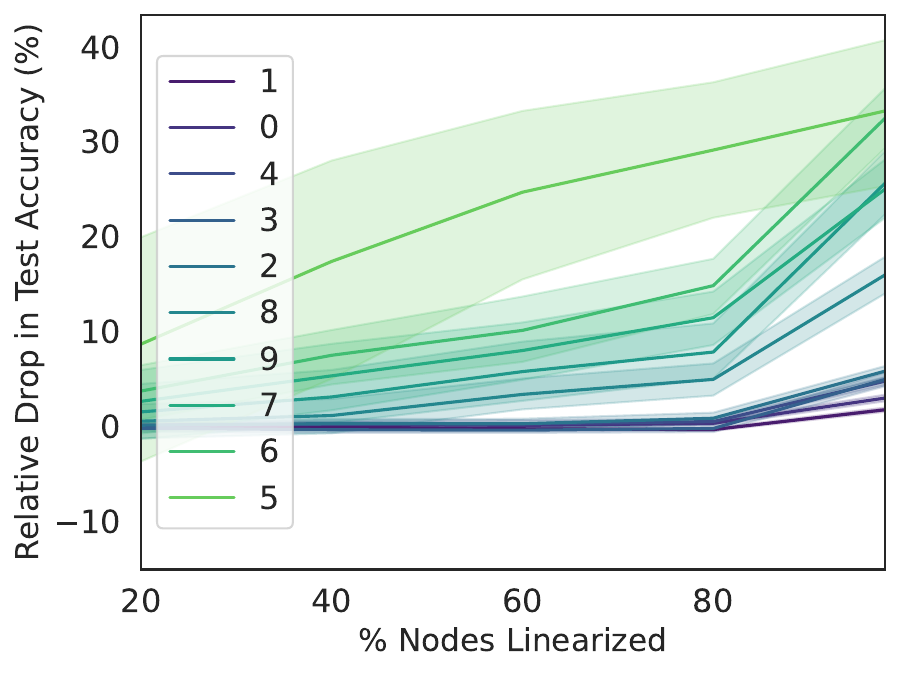}
    \includegraphics[width=0.24\textwidth]{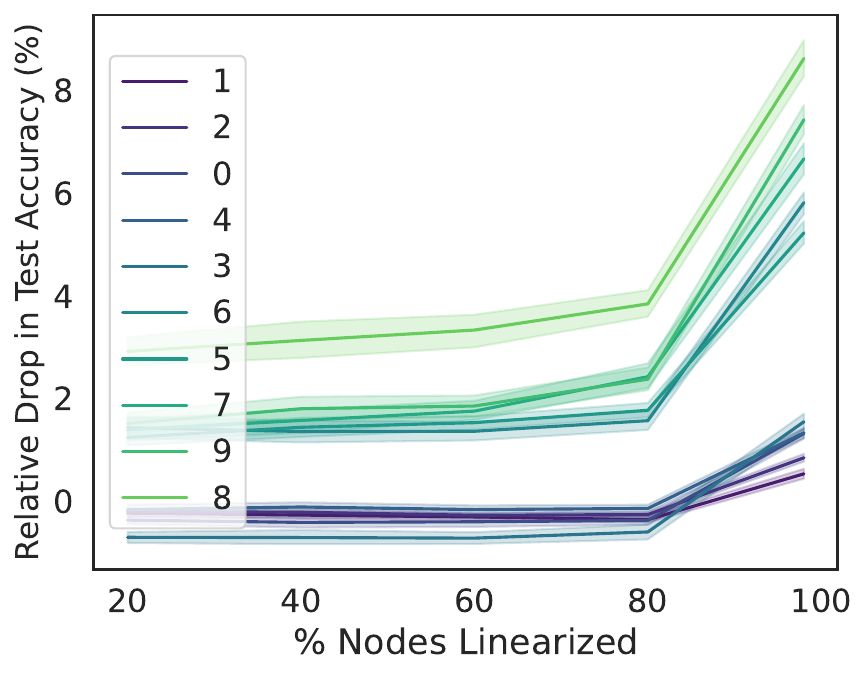}
    \includegraphics[width=0.24\textwidth]{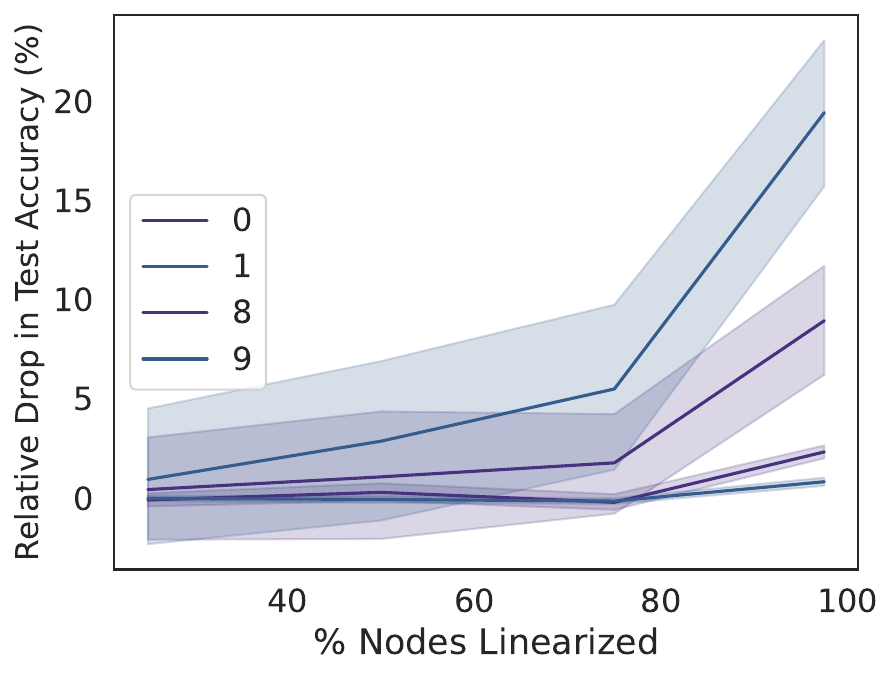}
    \includegraphics[width=0.24\textwidth]{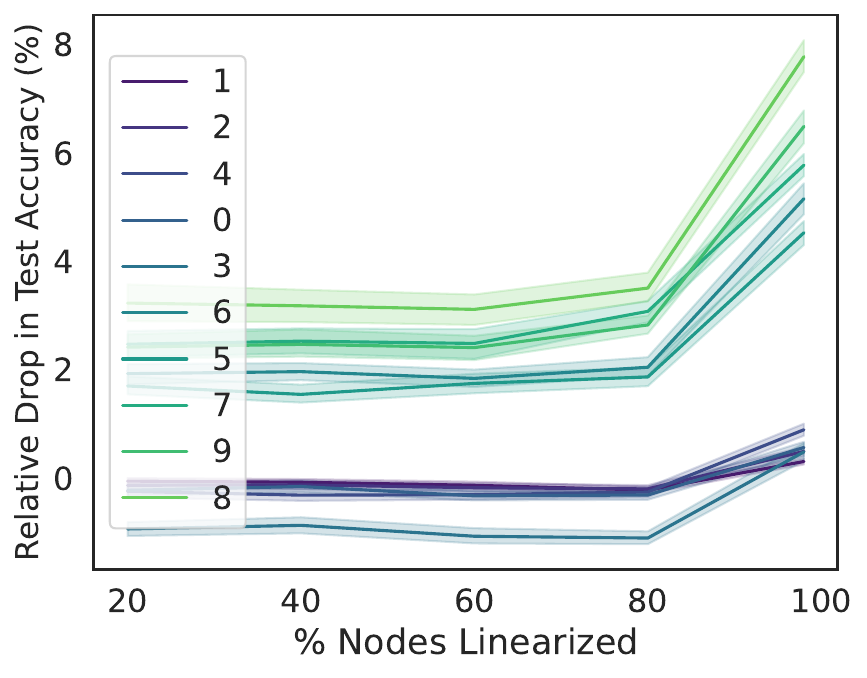}
    \caption{\gls*{snl}: For ResNet18 on CIFAR-10 (left) and SVHN (center left), and for ResNet34 on CIFAR-10 (center right) and SVHN (right)}
    \label{fig:appendix_non_mitigation_accs}
\end{figure}

In this section, we present more plots for relative accuracy drops in \cref{fig:appendix_non_mitigation_accs} to show the disparate impact of linearizing \gls*{relu}s. Observe again how minority groups suffer degradation in accuracy that is higher than for majority groups as the \gls*{relu} budget decreases/more \gls*{relu}s are linearized.

\section{Mitigation}
This subsection provides further evidence for the efficacy of the mitigation method discussed in \cref{sec:mitigation}. The labels in the legend refer to group indices.

\begin{figure}
    \centering
    \includegraphics[width=0.24\textwidth]{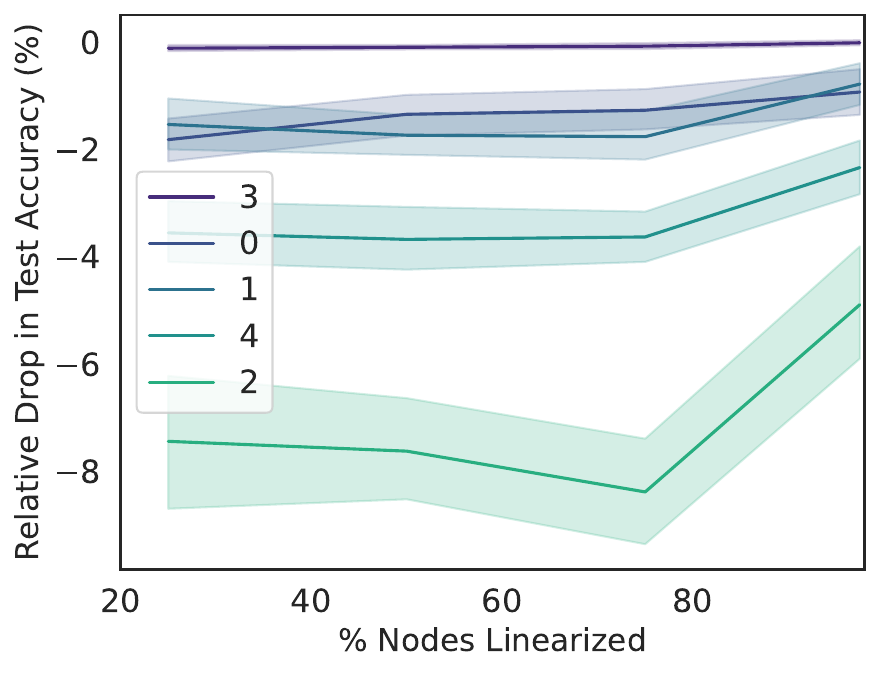}
    \includegraphics[width=0.24\textwidth]{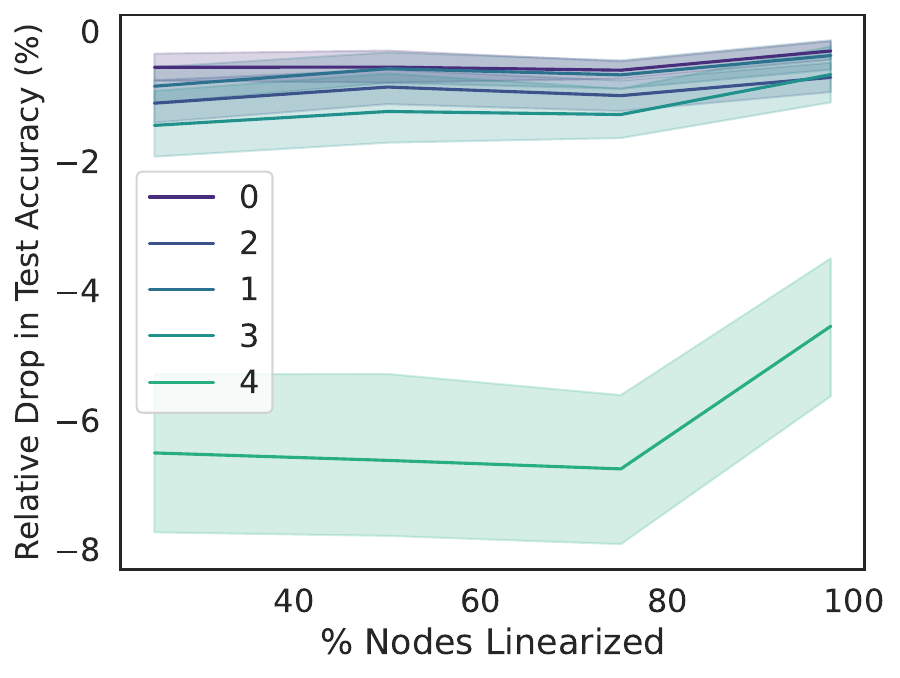}
    \includegraphics[width=0.24\textwidth]{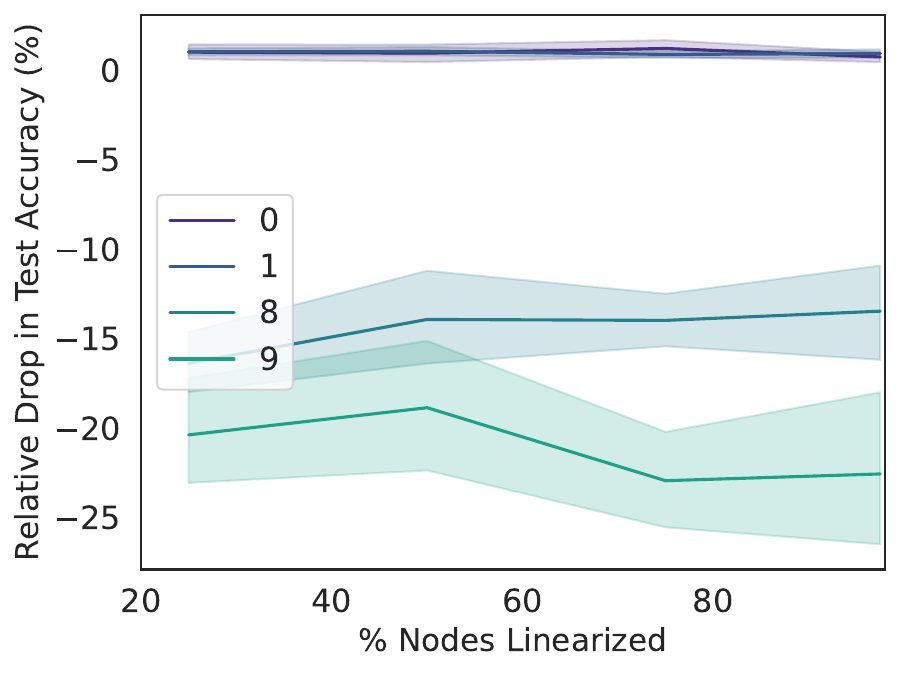}
    \includegraphics[width=0.24\textwidth]{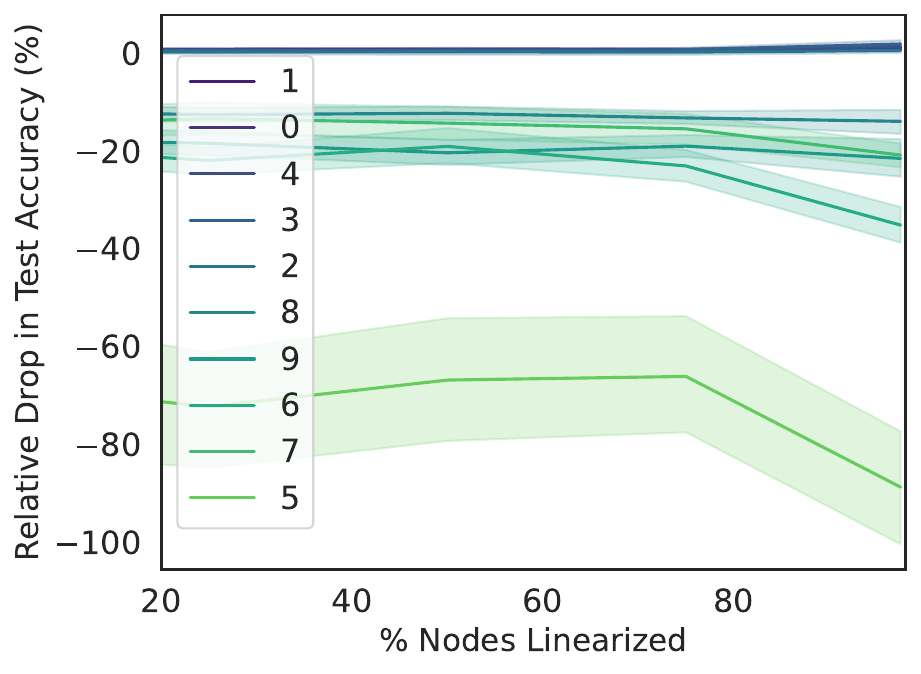}
    \caption{\gls*{snl} with Mitigation: For ResNet34 trained on UTKFace with age (left) and race (center left) labels and on CIFAR-10 (center right) and ResNet18 trained on CIFAR-10 (right).}
    \label{fig:appendix_mitigation}
\end{figure}



\subsection{Mitigation: Gradient Norms and Distance to the Decision Boundary}
\label{sec:app_mitigation_gradnorms}

\begin{figure}[!h]
    \centering
    \includegraphics[width=0.45\textwidth]{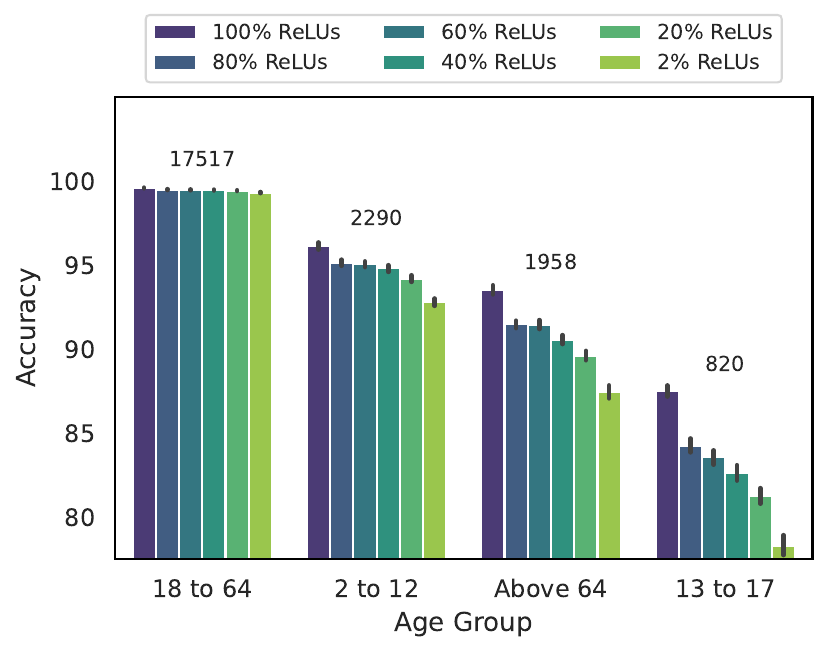}
    \includegraphics[width=0.45\textwidth]{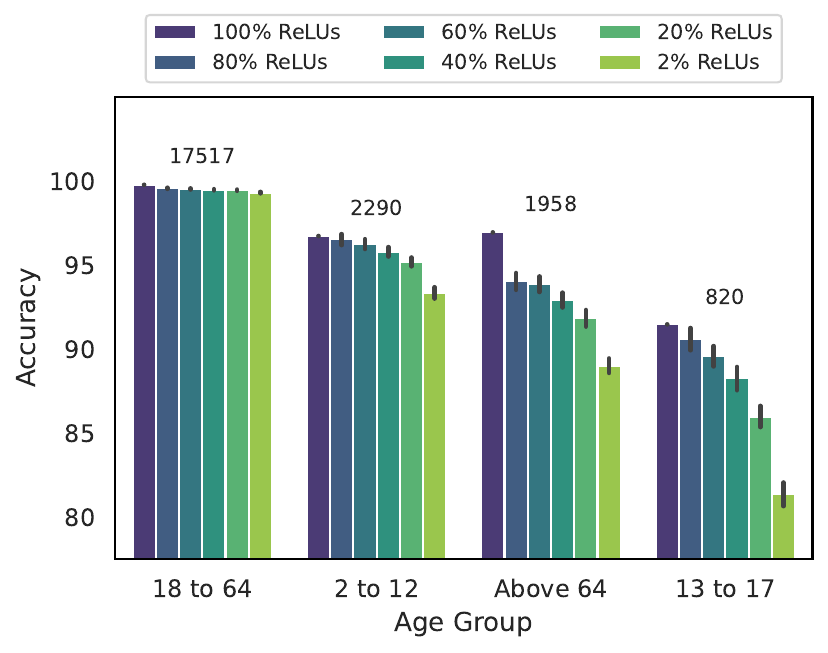}
    \caption{SNL: (Test) Accuracies for ResNet-18 trained on UTKFace with age labels; \textbf{left:} without mitigation; \textbf{right:} with mitigation.}
    \label{fig:Appendix_UTKFace_age_SNL_R18_BeforeAfter_GradNorms}
\end{figure}

\begin{figure}[!h]
    \centering
    \includegraphics[width=0.45\textwidth]{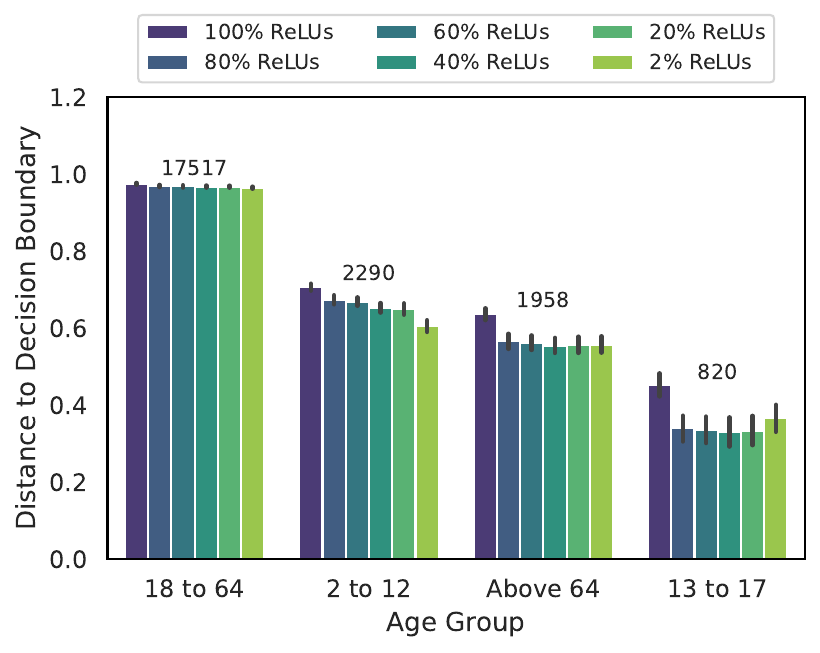}
    \includegraphics[width=0.45\textwidth]{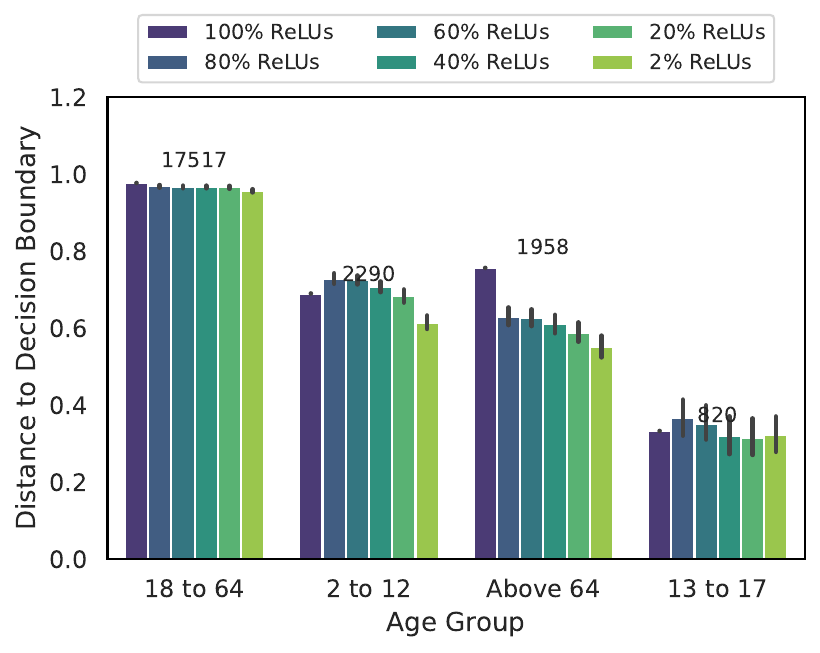}
    \caption{SNL: (Normalized) Distances to Decision Boundary for ResNet-18 trained on UTKFace with age labels; \textbf{left:} without mitigation; \textbf{right:} with mitigation.}
    \label{fig:Appendix_UTKFace_age_SNL_R18_BeforeAfter_DecBdy}
\end{figure}
The plots in \cref{fig:Appendix_UTKFace_age_SNL_R18_BeforeAfter_GradNorms} and \cref{fig:Appendix_UTKFace_age_SNL_R18_BeforeAfter_DecBdy} illustrate the differences in the values of gradient norms and (normalized) distances to the decision boundary when linearizing \gls*{relu}s with and without reduction. As mentioned in \cref{sec:mitigation}, a reduction in the gradient norms and an increase in the distance to the decision boundary can be observed when using the mitigation method.
\clearpage
\subsection{Mitigation: an ablation study of the parameter $\mu$}
\label{sec:app_mitigation_mu_ablation}
\begin{figure}[h!]
    \centering
    \begin{subfigure}{0.31\textwidth}
        \includegraphics[width=0.9\linewidth]{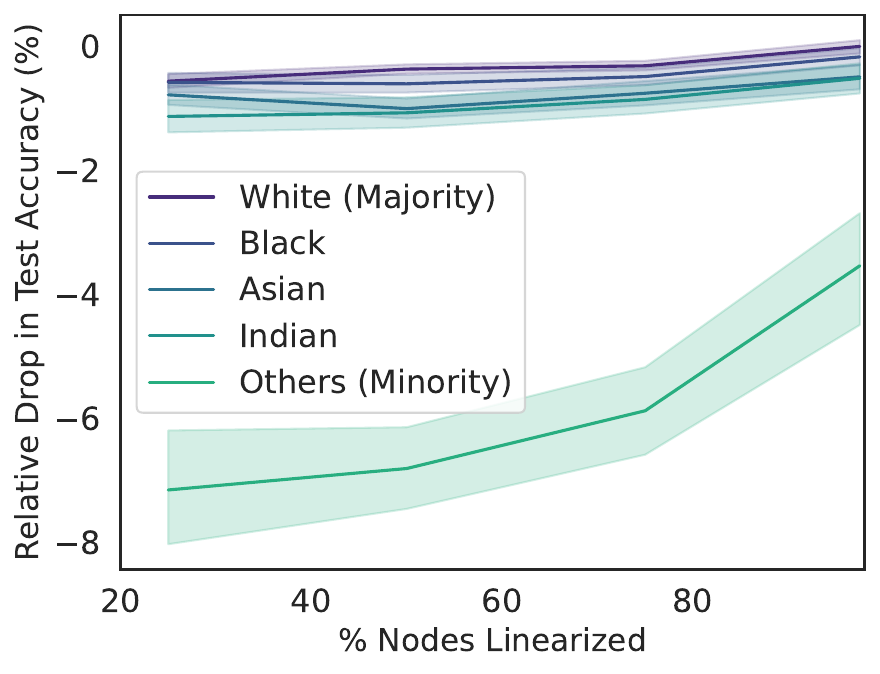}
        \caption{SNL on UTKFace race ResNet18 mitigation with $\mu=0.0005$}
        \label{fig:ablation_study_mu_1}
    \end{subfigure}
    \quad
    \begin{subfigure}{0.31\textwidth}
        \centering
        \includegraphics[width=0.9\linewidth]{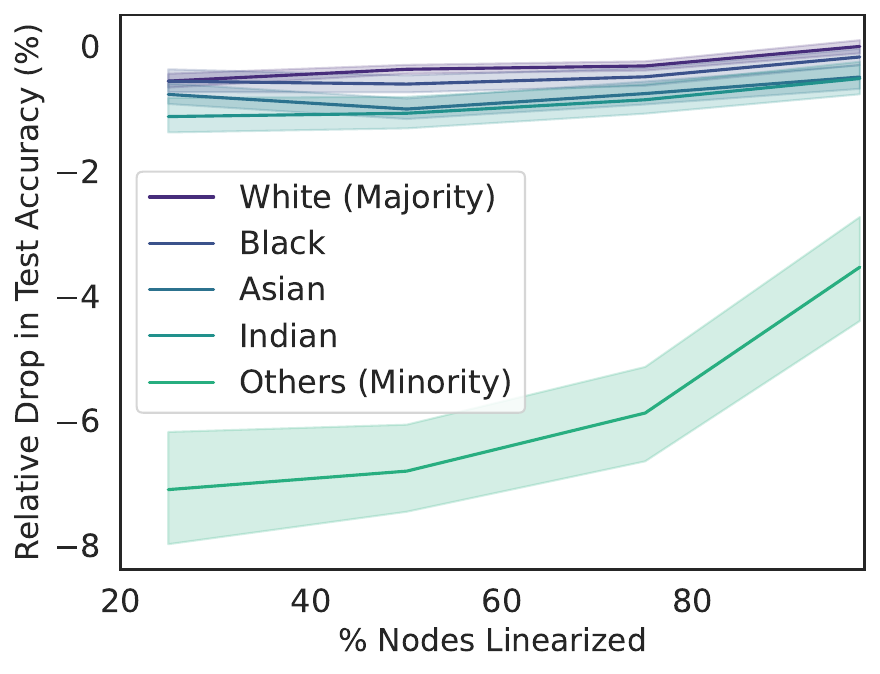}
        \caption{SNL on UTKFace race ResNet18 mitigation with $\mu=0.005$}
        \label{fig:ablation_study_mu_2}
    \end{subfigure}
    \quad
    \begin{subfigure}{0.31\textwidth}
        \centering
        \includegraphics[width=0.9\linewidth]{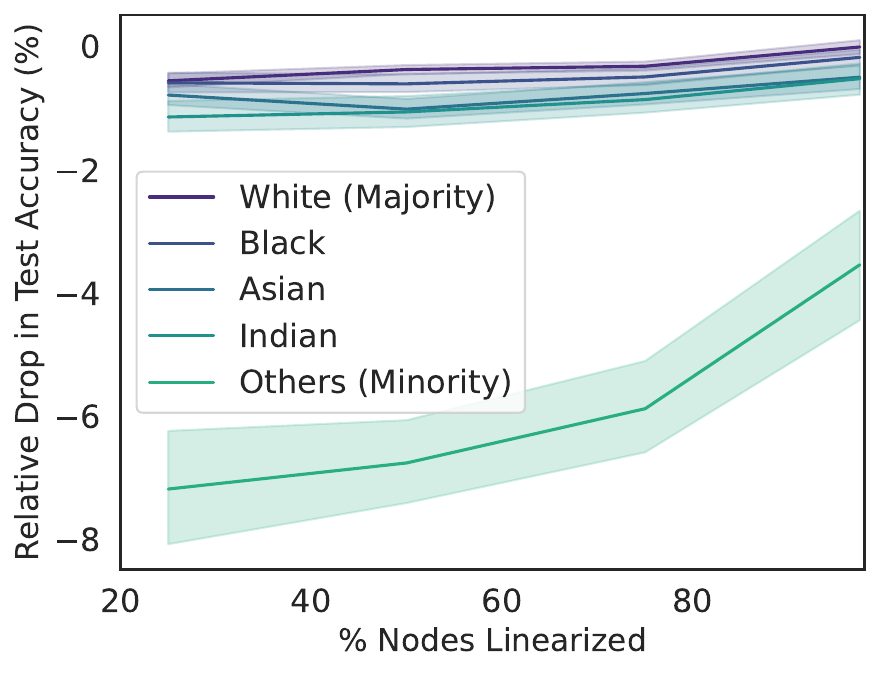}
        \caption{SNL on UTKFace race ResNet18 mitigation with $\mu=0.01$}
        \label{fig:ablation_study_mu_3}
    \end{subfigure}
\caption{We report the plots for the ablation study of the parameter $\mu$ for the SNL on UTKFace (race). We consider $\mu\in\{0.0005, 0.005,0.01\}$. In general we observe that these values of $\mu$ do not result in significant changes for the performance of the mitigation strategy. For the task at end, we could then say that there exists a neighborhood of values for $\mu$ where our mitigation consistently benefits the considered linearization techniques.}
\label{fig:ablation_study_mu}
\end{figure}

\end{document}